\documentclass[10pt,conference]{IEEEtran}
\IEEEoverridecommandlockouts
\usepackage{cite}
\usepackage{amsmath,amssymb,amsfonts}
\usepackage{algorithmic}
\usepackage{graphicx}
\usepackage{textcomp}
\usepackage{xcolor}
\usepackage{booktabs}

\def\BibTeX{{\rm B\kern-.05em{\sc i\kern-.025em b}\kern-.08em
    T\kern-.1667em\lower.7ex\hbox{E}\kern-.125emX}}
\begin{document}

\title{An Overview and Discussion of the Suitability of Existing Speech Datasets to Train Machine Learning Models for Collective Problem Solving}

\author{\IEEEauthorblockN{Gnaneswar Villuri}
\IEEEauthorblockA{\textit{Department of Electrical and Computer Engineering} \\
\textit{Stony Brook University}\\
Stony Brook, NY \\
gnaneswar.villuri@stonybrook.edu}
\and
\IEEEauthorblockN{Alex Doboli}
\IEEEauthorblockA{\textit{Department of Electrical and Computer Engineering} \\
\textit{Stony Brook University}\\
Stony Brook, NY  \\
alex.doboli@stonybrook.edu}
}

\maketitle

\begin{abstract}
This report characterized the suitability of existing datasets for devising new Machine Learning models, decision making methods, and analysis algorithms to improve Collaborative Problem Solving and then enumerated requirements for future datasets to be devised. Problem solving was assumed to be performed in teams of about three, four members, which talked to each other. A dataset consists of the speech recordings of such teams. The characterization methodology was based on metrics that capture cognitive, social, and emotional activities and situations. The report presented the analysis of a large group of datasets developed for Spoken Language Understanding, a research area with some similarity to Collaborative Problem Solving. 
\end{abstract}


\section{Introduction}

Our goal has been to study the process of problem solving by individuals, e.g., Individual Problem Solving (IPS), and teams, i.e., Collaborative Problem Solving (CPS)~\cite{Duke2022a, Villuri2024}. A main feature of our work has been studying team dynamics during CPS, as it is expressed by speech dialog during problem solving. In addition to studying Machine Learning (ML) and decision-making algorithms that use data collected from experimental settings~\cite{Villuri2024}, our work considered agent-based simulation models that mimic lab experiments~\cite{Doboli2021, Doboli2023, Doboli2019}, modeling of the behavior of research teams~\cite{Liu2022, Curiac2022}, and applications beyond problem solving, such as therapy~\cite{Doboli2022}. This previous work has highlighted the importance of having access to datasets that comprehensively represent the characteristics of IPS and CPS, so that the datasets can be used in ML training, for model construction, and analysis.     

The CPS process can be analyzed along four different perspectives: (i)~As a complex process that involves cognitive, social, and emotional aspects interrelated in many ways~\cite{Fiore2010, Sun2020, Wiese2019, Wiltshire2018}, (ii)~As a  process decided by the features of the tackled problems, which can pertain to three categories: well-defined, ill-defined, and open-ended problems~\cite{Schraw1995, Doboli2015, Doboli2014}, in which problems descriptions and CPS processes might incorporate constraints and requirements, like timing constraints, cost, robustness, ethical aspects, and so on~\cite{Medeiros2014, Forster2004}, (iii)~As a process that creates besides a solution also other outcomes, like learning new knowledge~\cite{Ellis2003} or new team behavior, i.e. how to effectively work in a team~\cite{Dillenbourg1999, Edmondson1999}, and (iv)~As a context-dependent process, in which the environment in which the process is performed is essential, such as organization, community, and society~\cite{Dodds2002, West2008}. Our research has focused mainly on understanding the interplay between cognitive, social, and emotional aspects during CPS.  

Using our research experience as a starting point, the goal of this report was to discuss the dataset requirements as well as the characteristics of the existing datasets in devising and training modern ML models, decision making methods, and analysis algorithms, so that they can improve the CPS process. This work assumed that CPS was conducted in small teams of around four members, which talk to each other while solving a problem together. A dataset represents the speech recordings of such teams. It is accepted that the quality of datasets is critical for ML training, e.g., the training and re-training of deep neural networks, like transformers and Large Language Models (LLMs).  The characterization methodology is based on the cognitive, social, and emotional activities of problem solving, which the report argues should be captured and characterized to understand any CPS process. The report enumerates these activities and summarizes the CPS process. Then, the report presentation refers to Spoken Language Understanding (SLU), a research domain in which spoken language datasets were devised to automatically model individual and team activities. We felt that SLU tasks resemble CPS activities to some level. A set of CPS-based metrics were proposed and used to characterize the main SLU datasets. Then, the report discusses the needs related to dataset development to devise new ML methods to improve the CPS process.


The report uses the term {\em utterance} to denote a unit of speech or text that represents a complete expression, such as a sentence, phrase, or command. It is typically defined by natural boundaries, such as pauses or shifts in conversational turns.

The report has the following structure. Section~II offers an overview of individual and collective problem solving. Section~III summarizes SLU and compares it to CPS. Section~IV offers a taxonomy of the datasets used in SLU, and the SLU activities that they address. Section~V presents the quantitative characterization to study the suitability of SLU datasets to be used to represent CPS. Section~VI discusses the CPS-based requirements of new datasets to be created. Conclusions end the report. 

\section {Overview of Individual and Collective Problem Solving}








\subsection {Problem Solving: Cognitive Dimension}

Problem-solving activities belong to the following categories~\cite{Sun2020, Wiese2019, Nijstad2006}:
\begin{enumerate}
\item
{\bf Problem Framing (PF)}: PF requires collecting the requirements and constraints that define a problem to produce a description of the problem-solving goals~\cite{Suk2021, Levin1998}. Goals can refer to functionality (e.g., the nature of the input processing) and performance (i.e. the processing speed). Descriptions might include ambiguities, unknowns, and can be only partially defined. Multiple problem framings are possible for a problem, especially for ill-defined and open-ended problems~\cite{Levin1998}. Reframing might be needed if the current description is incorrect or does not lead towards a solution~\cite{Gonzalez2005, Wirfs2006}.  

\item
{\bf Problem Understanding (PU)}: PU is the activity of associating meaning to a problem description~\cite{Brennan2010}. It involves identifying the inputs and outputs of the problem, their attributes (features), the connections between inputs and outputs, the variables involved in processing and their characteristics, the relationships between different input, output and variable connections, and the nature of the ambiguities and unknowns of a description.  

\item
{\bf Information Recollection (IR)}: IR involves retrieving from the memory the knowledge related to solving a problem~\cite{Anderson2000}. It involves information cueing from the memory and remembering similar problems, which were previously solved, and which can serve me modified or as analogies to solving the new problem~\cite{Spellman2006, Goel1997}. More complex situations involve knowledge restructuring and sudden insight~\cite{Reynolds2006}. 

\item 
{\bf Identifying the Solving Approach (ISA)}: ISA represents the activity of devising the general strategy for solving a problem~\cite{Dunbar1997, Hart2017}. It produces a description, which while cannot be executed as it is not detailed enough, it serves as a starting point for solution elaboration, so that the general strategy can be converted into an executable solution. The description describes the broad operations (processing steps) required to realize the problem description.  

\item
{\bf Problem Decomposition (Divide and Conquer) (DC)}: DC (segmentation) decomposes a problem into its subproblems and assigns subgoals to each subproblem~\cite{Smith1985}. The solution to the subproblems must be combined into the overall solution.  

\item
{\bf Solution Elaboration (SE)}: SE details a solution description deemed to be still unexecutable by adding details, like new variables and computations on the variables~\cite{Visser1991, Bonnardel2003}. The involved approaches for elaboration include trail-and-error, adjustment and adaptation of existing fragments, combining solution fragments~\cite{Doboli2014, Costello2000, Wisniewski1998}, and mapping onto previous solutions (analogy)~\cite{Rietzschel2014}.  

\item 
{\bf Solution Analysis (SA):} SA finds the errors of a solution as well as its pros and cons as compared to alternative solutions~\cite{DeCaro2007, Deo2024}. 

\item
{\bf Solution Modification and Correction (SMC):} SMC involves modifying a solution to improve its performance or correct its errors~\cite{Regenwetter2024, Chilton2021}. This activity requires understanding the context in which the modification must be made, e.g., which solution fragments will work together with the modification, as well as devising a set of modification options. 

\item 
{\bf Addressing Fixation (AF)}: AF tackles cases in which problem solving is stuck, either by repeating previous solution steps or not being able to continue the solving process~\cite{Storm2015}. Typical AF activities include brainstorming or identifying new cues to help to unstuck the solving process~\cite{Paulus2006}.  

\item
{\bf Solution Reuse (SR)}: SR requires devising a solution, so that most of its fragments (components) can be reused in similar situations. SR is popular in repetitive problem solving situations, like devising a new version of an existing solution, in which the new problem only incrementally changes the requirements. Typical solution include devising component libraries, patterns~\cite{Wirfs2006}, templates~\cite{Doboli2014, Wei2007, Tang2006}, which while not strictly needed for creating a solution, they are useful in creating a sequence of related solution by reusing existing solution fragments.  

\item 
{\bf Knowledge Learning and Restructuring (KLR)}: KLR represents the learning of new knowledge and creating new associations by the solver during the problem solving process as well as any restructuring of the existing knowledge as a result of new insight from solving a new problem~\cite{Chi1981}. For example, a certain solution fragment might reused to address a new requirement. KLR also includes goal adjustment, belief change, and preference and priority adjustments. 
\end{enumerate}

\subsection {Problem Solving: Social Dimension}

CPS includes the following social activities that are not part of SLU:
\begin{enumerate}
\item 
{\bf Social Understanding (SU):} SU indicates the degree to which a team member creates an image of the other team members~\cite{Cacioppo2006}. This image includes cognitive, social, and emotional aspects, and is used during problem solving.

\item
{\bf Bridging Knowledge Gaps (BKG):} BKG is the process of addressing the knowledge gaps between the members of a team. This involves aspects, like the psychological safety~\cite{Edmondson1999} to feel comfortable asking questions and giving feedback, degree of participation in the process, and providing answers to posed questions.  

\item
{\bf Team Agreement (TA):} TA presents the knowledge on which all team members agree on~\cite{Cronin2019, Vesper2016}. 

\item
{\bf Team Synchronization (TS):} TS represents the degree to which team members participate and identify social interaction procedures that support effective team coordination to solve a problem~\cite{Cronin2019}.  

\item
{\bf Individual Learning (IL):} IL describes the amount of an individual's learning by participating to CPS, including new knowledge learned as well as SU~\cite{Perry2006}. While KLR is in a solitary situation, IL refers to individual learning in a social context. Traditionally, IL includes goal adjustment, new knowledge retainment, creating new associations~\cite{Brown2002}, belief change, insight~\cite{Reynolds2006}, pressure management, and developing team management skills. 

\item
{\bf Team Learning (TL):} While BKG, TA, TS are sometimes considered part of TL, this work associates activities to TL, like goal sharing, role distribution, self-construal orientation, and conflict resolution~\cite{Edmondson1999, Edmondson2004}.   
\end{enumerate}

\subsection {Problem Solving: Emotional Dimension}

IPS and CPS includes the following aspects related to emotions, and which are not part of SLU:
\begin {enumerate}
\item 
{\bf Individual Emotional Behavior (IEB)}: IEB refers to emotion, motivation, self-efficacy, pressure management, evaluation apprehension, and cultural characteristics~\cite{Girardi2020, Barth2010, Ford2015}. 

\item
{\bf Team Emotional Behavior (TEB)}: TEM refers to psychological safety~\cite{Edmondson1999}.  
\end {enumerate}

\section {Related Research Areas}

\subsection{Activities during Spoken Language Understanding}

SLU systems perform the following main activities, each of which contributes to robust language understanding:
\begin{enumerate}
\item 
\textbf{Speech-to-Text Understanding (STU):} STU converts spoken input into text, which is then processed for downstream tasks~\cite{googlecloud}.

\item \textbf{Domain Classification (DC):} DC assigns utterances to predefined categories depending on the application domain, like weather, music, or travel~\cite{taus}.

\item 
\textbf{Intent Recognition(IR):} IR determines the purpose or action implied by an utterance~\cite{gfgintent}. For example, in the sentence ``{\tt \small Set a timer for 10 minutes},'' the communicated intent is \texttt{\small SetTimer}.

\item \textbf{Named Entity Recognition (NER):} NER identifies and classifies entities, such as dates, locations, or product names~\cite{ner}.

\item \textbf{Slot Filling (SF):} SF extracts specific parameters required to fulfill the identified intent~\cite{sf}. In the above example, the value \texttt{\small 10 minutes} is extracted as the slot value for the parameter \texttt{\small duration}.

\item \textbf{Dialogue State Tracking (DST):} DST maintains the conversational context by storing user goals, preferences, and prior interactions~\cite{dst}.
    
\item \textbf{Question Answering (QA):} QA answers queries, like ``{\tt \small What is the weather in New York?}'' by extracting relevant information from structured or unstructured data sources~\cite{qa}.

\item \textbf{Action Prediction (AP):} AP predicts appropriate responses or actions, such as confirming details or initiating database queries~\cite{ap}.
\end{enumerate} 
This work analyzed activities~(2)-(8) to discuss the similarities between IPS, CPS, and SLU. Activity~1 (STU) is addressed in other work, e.g.,~\cite{stu}.

\subsection {Discussions of the Computational Activities to Support IPS and CPS}

The following differences distinguish IPS and CPS activities from the SLU activities:

\begin {enumerate}
\item
{\bf Multi-modal tracking:} The data used in IPS and CPS is multi-modal, not only speech data converted into text but also outcomes of the solving process, like designs, code, etc., physiological signals (e.g., heart rate, sweat, etc.), eye gaze, body posture, and so on~\cite{Pelachaud2005, Ciceri2006}. Multi-data collection, integration, and fusion into a coherent representation are required for team tracking and characterization.

\item
{\bf Semantic parsing and understanding:} IPS and CPS require the identification of ten different types of activities, i.e. activities~(2)-(11) in Section~II.C, while SLU includes mainly two activities, IR and SF. In contrast to SLU, IPS and CPS must bridge between different description styles, e.g., using sequences of conditions on data features, sequences of processing steps, or function compositions~\cite{Villuri2024, Villuri2025a, Villuri2025b}. Connecting related entities at different abstraction levels is difficult as the connections might not be always reducible to the expressed features. A representation is created as part of understanding and then used in solving. Descriptions are of different complexities (e.g., mixtures of description styles) and sizes, and at different levels of abstraction, and can include ambiguities, unknowns, and partial solutions. Semantic parsing must address these issues.

\item 
{\bf Specific activities:} IPS and CPS involve specific activities, which arguably are not present in SLU, like PF and PU. Both activities can produce different representations of the problem requirements, e.g., the priorities associated to the requirements. CPS needs reaching a consensus about the different representations. Moreover, handling ambiguities, unknowns, and incomplete problem descriptions.    

\item
{\bf Solution elaboration}: The problem solving process involves the elaboration of the description in contrast to SLU activities in Section~III.A, which do not require elaboration. Elaboration requires insertion of new parameterized structures and connecting the parameters through associations and causality relations. Moreover, problem solving repeatedly performs DC, SA, SMC, and SR, which are less important in dialog systems, in which the objectives of a conversation, e.g., getting correct answers to precise questions.   

\item
{\bf Reactivity to unexpected situations:} The activities refer to handling of unexpected situations, e.g., situations that cannot be predicted based on previous experiences, including the detection and management of such situations. Broadly, it requires effective out-of-the-box thinking, including trial-and-error to learn new insight, systematic formulation of hypotheses, and running experiments to check them.     

\item
{\bf Social and emotional feature management:} Processing social and emotional features is important in CPS but arguably less important for SLU. It involves the items enumerated in Sections~II.D and II.E, and how they influence the problem solving process.  

\item
{\bf Understanding the issues of an individual working in a team:} CPS requires understanding the cognitive, social, and emotional impact created on an individual by his/her participation to the teamwork. This involves IL and IEB.

\item
{\bf Problem solving process:} In contrast to SLU, which usually involves a small number of activities, like giving a response to a question, IPS and CPS require tracking the problem solving process over time to understand the characteristics of the process, like the degree to which there is progress towards finding a solution, reaching a consensus about the solution, identifying solution alternatives, analyzing the pros and cons of solutions, ending-up in unexpected situations, and learning by participating agents. However, the expected final outcomes are less well defined, as solving might not always produce a solution. Precise identification and characterization of the IPS and CPS trace over time are critical to understand the connections between activities, as opposed to DST for SLU, which arguably is less intensive as SLU activities are more transactional. The IPS and CPS is more ad-hoc, less systematic as it can switch among different kind of activities without following a precise pattern. AF is also critical in problem solving but less likely in SLU. 

\end {enumerate}





\section{Taxonomy of SLU Datasets}

\begin{figure*}[htbp]
    \centering
    \includegraphics[width=\textwidth, angle=0]{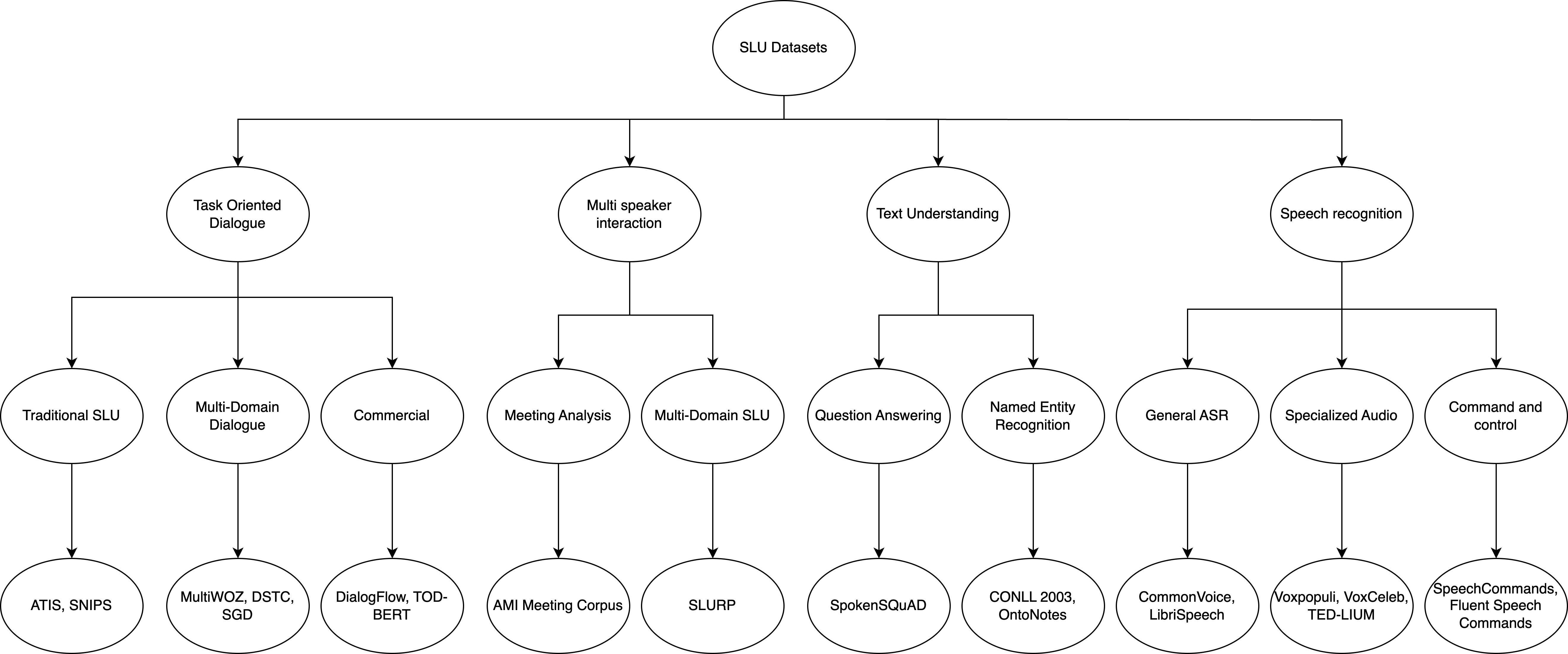}
\caption{SLU Datasets Classification}
    \label{fig:slu-taxonomy}
    \vspace * {-0.2in}

\end{figure*}

Figure~\ref{fig:slu-taxonomy} presents a \textbf{taxonomy of Spoken Language Understanding (SLU) datasets} organized by their primary purpose. The datasets are grouped into four major categories based on their application areas: \textbf{Task-Oriented Dialogue
}, \textbf{Multi-Speaker Interaction}, \textbf{Text Understanding}, and \textbf{Speech Recognition}. Each category is further divided into subcategories, highlighting the specific use cases of the datasets, with examples provided under each subcategory.

\begin{itemize}
    \item \textbf{Task-Oriented Dialogue} \\
    These datasets are designed to evaluate the performance of Spoken Language Understanding (SLU) algorithms, focusing on tasks such as intent detection, slot filling, and dialogue state tracking. They are typically used in task-specific scenarios, such as air ticket reservations~\cite{atis1990}, or in multi-domain contexts like voice assistant commands~\cite{multiwoz2018}. 
    The subcategories are: Traditional SLU, which includes single-domain datasets such as ATIS~\cite{atis1990} and SNIPS~\cite{snips2018} that focus on simple, constrained tasks; Multi-Domain Dialogue, which includes datasets covering multiple domains, such as MultiWOZ~\cite{multiwoz2018}, DSTC~\cite{dstc2014}, and SGD~\cite{sgd2020}, requiring models to handle complex domain transitions; and Commercial Datasets, which are proprietary datasets like DialogFlow~\cite{dialogflow} and TOD-BERT~\cite{todbert2020}, often used in industry and incorporating real-world applications.

    \item \textbf{Multi-Speaker Interaction} \\
    These datasets focus on conversations involving multiple participants, often in collaborative settings. For example, the AMI Meeting Corpus~\cite{amimeeting2005} contains recordings of business meetings annotated for actions, dialogue acts, and participant roles. Unlike task-oriented dialogue datasets, these focus on analyzing group dynamics, interactions, and the roles of speakers in conversations.

    \item \textbf{Text Understanding} \\
    These datasets aim to improve understanding of spoken or written queries, including tasks like entity extraction and question answering. Multi-Domain SLU datasets such as SLURP~\cite{slurp2020} focus on understanding spoken language across diverse domains, emphasizing the challenge of generalization. Question Answering datasets like SpokenSQuAD~\cite{spokensquad2018} adapt the popular SQuAD dataset to spoken input, focusing on processing natural spoken queries. Named Entity Recognition (NER) datasets, such as CONLL 2003~\cite{conll2003} and OntoNotes~\cite{ontonotes2013}, specialize in identifying entities like names of people, organizations, and locations within text, offering precision-focused evaluation for NER models.

    \item \textbf{Speech Recognition} \\
    These datasets support tasks like automatic speech recognition (ASR), speaker identification, and command recognition. General ASR datasets, such as CommonVoice~\cite{commonvoice2020} and LibriSpeech~\cite{librispeech2015}, provide diverse data across languages and accents to ensure generalizability. Specialized Audio datasets, such as VoxPopuli~\cite{voxpopuli2021}, VoxCeleb~\cite{voxceleb2017}, and TED-LIUM~\cite{tedlium2014}, cater to specific applications like speaker verification or lecture-style speech transcription. Command and Control datasets, such as SpeechCommands~\cite{speechcommands2018} and Fluent Speech Commands~\cite{fsc2019}, focus on specific spoken commands used in smart devices, emphasizing clarity and conciseness.
\end{itemize}

\begin{table*}
\caption{Comparative Analysis of Spoken Language Understanding Datasets}
\label{tab:slu-comparison}
\renewcommand{\arraystretch}{1.3}
\begin{tabular}{@{}p{2cm}p{2.5cm}p{2.5cm}p{3cm}p{3cm}p{2.5cm}@{}}
\toprule
\textbf{Dataset} & \textbf{Size} & \textbf{Purpose} & \textbf{Data Collection} & \textbf{Sentence Characteristics} & \textbf{Annotations} \\
\midrule
ATIS & 6300 utterances & Airline reservation systems & Simulated user queries & Short (5–10 words), concise queries & Intent labels, slot annotations \\
\midrule
SNIPS & 14,000 utterances across 7 domains & General-purpose voice assistants & Crowdsourced utterances & Medium (10–20 words), moderately detailed & Intent labels, slot-value pairs \\
\midrule
Frames & 1,369 utterances & Task-oriented dialogue systems & Wizard-of-Oz experiments & Variable length (5–30 words), diverse travel tasks & Dialogue-level annotations \\
\midrule
DSTC & ~24,000 utterances (DSTC2) & Dialogue state tracking & Real-world dialogues & Medium-length (10–20 words) utterances & Slot-value pairs, system states \\
\midrule
MultiWOZ & 115,000+ utterances & Multi-domain dialogue & Wizard-of-Oz setup & Multi-turn conversations (variable, 10–50 words) & Rich dialogue-level annotations \\
\midrule
TOD-BERT & 100,000+ utterances & Model pretraining & Aggregated datasets & Domain-specific, task-oriented, medium (10–20 words) & Standardized cross-dataset annotations \\
\midrule
Google Dialogflow & Proprietary (size not publicly disclosed) & Commercial spoken language understanding & Real-world user queries & Diverse, multi-domain utterances (5–50 words) & Intent and entity annotations \\
\midrule
SGD (Schema-Guided Dialogue) & 20,000+ utterances, 300+ services & Dialogue state tracking and intent detection & Simulated user queries & Multi-turn conversations (10–50 words) & Intent labels, slot-value pairs \\
\midrule
CONLL (2003) & 22,137 utterances & Named entity recognition (NER) & News articles & Well-formed, formal text (15–25 words) & Entity labels (PER, LOC, ORG, etc.) \\
\midrule
OntoNotes & 1.5M words & Coreference resolution, NER, and more & Multi-source (news, conversational speech, etc.) & Formal and conversational text (10–30 words) & Rich multi-layered annotations \\
\midrule
LibriSpeech & 1,000 hours of audio & Automatic speech recognition (ASR) & Audiobooks & Long, formal sentences (30–50 words) & Word-level transcriptions \\
\midrule
Common Voice & 13,000+ hours of audio & ASR and TTS (text-to-speech) training & Crowdsourced recordings & Wide variety of text (5–25 words) & Word-level transcriptions \\
\midrule
SpokenSQuAD & 100,000+ questions & Spoken question answering & Text-to-speech applied to SQuAD dataset & Well-formed questions (10–20 words) & Question-answer pairs \\
\midrule
Natural Questions & 300,000+ questions & Open-domain QA & Real user queries from Google Search & Diverse, natural queries (5–25 words) & Long-answer and short-answer annotations \\
\midrule
Fluent Speech Commands (FSC) & 30,000+ utterances & Voice-controlled smart devices & Simulated command-like utterances & Short, directive sentences (5–10 words) & Intent labels \\
\midrule
VoxCeleb & 1,000+ hours of audio & Speaker identification and verification & Interview recordings & Natural noise, multi-speaker, variable length (10–30 words) & Speaker IDs \\
\midrule
LibriLight & Subset of LibriSpeech & Low-resource ASR and SLU tasks & Curated audiobooks & Long, formal sentences (30–50 words) & Word-level transcriptions \\
\midrule
SLURP & ~72,000 utterances & Multi-domain SLU tasks & Real-world scenarios & Diverse, task-oriented (10–25 words) & Intent labels, entity extraction \\
\midrule
Speech Commands & 10,000+ utterances & Keyword spotting and command recognition & Short commands & Single-word utterances (1 word) & Command labels \\
\midrule
VoxPopuli & 400,000+ hours of audio & Multilingual SLU tasks & Multilingual recordings & Diverse languages, variable (5–50 words) & Transcriptions, speaker IDs \\
\midrule
TED-LIUM & 452+ hours of audio & ASR and SLU & TED talks & Long, formal speeches (30–50 words) & Transcriptions, speaker labels \\
\midrule
AMI Meeting Corpus & ~100 hours of audio & SLU in collaborative settings & Meeting scenarios & Multi-speaker dialogues (10–30 words) & Transcriptions, topics, discourse annotations \\
\bottomrule
\end{tabular}
\end{table*}

Table~\ref{tab:slu-comparison} provides a detailed comparison of widely used datasets relevant to Spoken Language Understanding (SLU). Each column in the table represents a specific aspect of these datasets, making it easier to evaluate their suitability for various research tasks.

\begin{itemize}
    \item \textbf{Dataset}: Name of the dataset.
    \item \textbf{Size}: Size of the dataset, measured in utterances, hours of audio, or words. 
    \item \textbf{Purpose}: The main goal or application area for which the dataset is designed, such as intent detection, dialogue systems, or speech recognition.
    \item \textbf{Data Collection}: The method used to gather data, such as real-world scenarios, Wizard-of-Oz experiments, or crowdsourced contributions.
    \item \textbf{Sentence Characteristics}: Specific features of the sentences or utterances in the dataset, including length, formality, and content diversity.
    \item \textbf{Annotations}: The types of labels or metadata available in the dataset, such as intents, slot-value pairs, or transcriptions.
\end{itemize}

\subsection*{Explanation of Each Dataset}

\begin{enumerate}
\item{\bf ATIS\cite{atis1990}:} Designed for airline reservation systems with 6,300 utterances. Sentences concise user queries. Annotated with intent labels (e.g., booking flights) and slot values (e.g., destination, date).
\item{\bf SNIPS\cite{snips2018}:} General-purpose dataset for voice assistants, containing 14,000 utterances across 7 domains. Utterances are of moderate length (10-20 words), suitable for intent classification. Includes intent labels and slot-value pairs.

\item{\bf Frames\cite{frames2017}:} Collected for task-oriented dialogue systems using Wizard-of-Oz experiments. Contains diverse travel tasks with variable sentence lengths. Provides dialogue-level annotations to capture interactions.

\item{\bf DSTC\cite{dstc2014}:} Focuses on dialogue state tracking, with about 24,000 utterances (DSTC2). Utterances are medium-length and derived from real-world dialogues. Annotated with slot-value pairs and system states.

\item{\bf MultiWOZ\cite{multiwoz2018}:} Large-scale dataset for multi-domain dialogue systems with over 115,000 utterances. Involves multi-turn conversations collected through Wizard-of-Oz experiments. Offers rich dialogue-level annotations for tracking intents and slots.

\item{\bf TOD-BERT\cite{todbert2020}:} Created for model pretraining with over 100,000 utterances aggregated from multiple sources. Focuses on domain-specific, task-oriented text. Provides standardized annotations for cross-dataset generalization.

\item{\bf Google Dialogflow\cite{dialogflow}:} A proprietary dataset designed for commercial SLU tasks. Derived from real-world user queries across diverse domains. Annotated with intents and entities for SLU tasks.

\item{\bf SGD (Schema-Guided Dialogue)\cite{sgd2020}:} Contains 20,000+ utterances across 300+ services for dialogue state tracking and intent detection. Simulated user interactions in multi-turn conversations. Provides intent labels and slot-value pairs.

\item{\bf CONLL (2003)\cite{conll2003}:} Classic dataset for named entity recognition (NER) tasks, with 22,137 utterances from news articles. Sentences are well-formed and formal. Annotated with entity types like PER (person), LOC (location), and ORG (organization).

\item{\bf OntoNotes\cite{ontonotes2013}:} Large-scale dataset with 1.5M words, used for tasks like coreference resolution and NER. Data comes from multi-sources (news, speech, and conversations). Provides multi-layered annotations for diverse NLP tasks.

\item{\bf LibriSpeech\cite{librispeech2015}:} Audio dataset with 1,000+ hours from audiobooks for automatic speech recognition (ASR). Sentences are long and formal. Includes word-level transcriptions.

\item{\bf Common Voice\cite{commonvoice2020}:} Crowdsourced audio dataset with 13,000+ hours for ASR and TTS tasks. Covers a wide variety of text and speakers. Annotated with word-level transcriptions.

\item{\bf SpokenSQuAD\cite{spokensquad2018}:} Derived from SQuAD by applying text-to-speech for spoken question answering. Contains over 100,000 well-formed questions. Provides question-answer pairs as annotations.

\item{\bf Natural Questions\cite{naturalquestions2019}:} Large dataset with 300,000+ queries from Google Search for open-domain QA. Queries are diverse and natural. Annotated with long-answer and short-answer labels.

\item{\bf Fluent Speech Commands (FSC)\cite{fsc2019}:} Contains over 30,000 utterances specifically crafted for voice-controlled smart devices. The sentences are concise and action-oriented (e.g., “turn on the lights”) and are accompanied by intent labels, making it suitable for training and evaluating systems designed for clear and directive command recognition.

\item{\bf VoxCeleb\cite{voxceleb2017}:} Audio dataset with 1,000+ hours from interviews for speaker identification tasks. Contains natural noise and multi-speaker interactions. Annotated with speaker IDs.

\item{\bf LibriLight\cite{librilight2020}:} Subset of LibriSpeech for low-resource ASR tasks. Contains long, formal sentences from curated audiobooks. Annotated with word-level transcriptions.

\item{\bf SLURP\cite{slurp2020}:} Multi-domain dataset with 72,000 utterances for SLU tasks. Reflects real-world scenarios with diverse sentence types. Provides intent labels and entity extraction.

\item{\bf Speech Commands\cite{speechcommands2018}:} Contains 10,000+ utterances for keyword spotting tasks. Features single-word utterances for command recognition. Annotated with command labels.

\item{\bf VoxPopuli\cite{voxpopuli2021}:} Multilingual dataset with 400,000+ hours of audio for SLU tasks. Contains multilingual recordings. Annotated with transcriptions and speaker IDs.

\item{\bf TED-LIUM\cite{tedlium2014}:} Audio dataset with 452+ hours of TED talk recordings. Sentences are long and formal, derived from speeches. Provides transcriptions and speaker labels.

\item{\bf AMI Meeting Corpus\cite{amimeeting2005}:} Dataset with ~100 hours of audio for SLU in meeting scenarios. Contains multi-speaker dialogues. Annotated with transcriptions, topics, and discourse-level annotations.

\end{enumerate}

\subsection{How Do Existing Data Sets Address SLU Activities}

Existing datasets have played a main role in advancing SLU research by providing annotated examples that mirror real-world interactions. We discussed next how the datasets have contributed to the study of the various SLU tasks:

\subsubsection{Intent Recognition (IR) and Slot Filling (SF)}

IR and SF are two key components in SLU, where the system must understand the user's intention and extract relevant information \cite{tur2011spoken}. Datasets, like \textbf{ATIS}~\cite{atis1990} and \textbf{Snips}~\cite{snips2018}, have been particularly popular in this regard. 

The \textbf{ATIS} dataset, focused on the travel domain, provides utterances that are labeled with intents such as \texttt{\small FlightSearch}, and slots like \texttt{\small origin\_city} and \texttt{\small destination\_city}. These annotations enable systems to identify the user's intent (e.g., booking a flight) and extract useful details from their input, making it easier to handle travel-related tasks.

The \textbf{Snips} dataset spans a broader range of domains, such as music, weather, and smart home control. It includes both intent labels and slot annotations, helping systems learn to recognize a wide variety of intents and extract information in different contexts. 

By providing these domain-specific examples, both datasets allow models to generalize across different areas, thus supporting the development of versatile SLU systems.

\subsubsection{Domain Classification (DC)}

DC is essential for SLU systems to manage different topics or services within a conversation \cite{mcTear2016}. Datasets like \textbf{Google Dialogflow}~\cite{dialogflow} and \textbf{SGD}~\cite{sgd2020} provide valuable resources for training models to handle multi-domain conversations. 

The \textbf{\small Google Dialogflow} dataset, for example, includes utterances annotated with domain tags that cover a diverse range of topics, from entertainment to healthcare. This helps systems identify which domain the user's query belongs to, ensuring that the correct response or action is taken.

Similarly, \textbf{SGD} (Spoken Language Understanding Dataset) offers a collection of dialogues with domain annotations, enabling systems to switch seamlessly between different topics. 

These multi-domain datasets are particularly useful for building virtual assistants and multi-functional chatbots that need to interpret and respond to a wide array of requests in a single conversation.

\subsubsection{Dialogue State Tracking (DST)}

DST is a critical task in multi-turn dialogues, where the system must keep track of the evolving conversation state \cite{xing2024dc}. Datasets like \textbf{MultiWOZ}~\cite{multiwoz2018} and \textbf{DSTC}~\cite{dstc2014} are designed to support this task by providing annotations that capture the dialogue state across several turns. 

\textbf{MultiWOZ}, for instance, is a large-scale dataset that covers domains such as hotel bookings, restaurant reservations, and public transportation. It annotates each dialogue with information about user intents, requested slots (e.g., \texttt{\small hotel\_name}, \texttt{\small check\_in\_date}), and system actions, helping systems manage the state of the conversation over multiple interactions.

Likewise, \textbf{DSTC} (Dialogue State Tracking Challenges) includes datasets that focus on tracking the dialogue state, such as user goals and system responses, which are crucial for ensuring that the system remains contextually aware throughout the conversation. 

These datasets help SLU models develop the capability to manage long, complex dialogues and ensure that responses are contextually appropriate.

\subsubsection{Named Entity Recognition (NER)}

NER is an essential SLU task that involves identifying and classifying named entities (such as names, locations, and dates) in a given text \cite{altexsoft2023ner}. Datasets like \textbf{CoNLL-2003}~\cite{conll2003} and \textbf{OntoNotes}~\cite{ontonotes2013} serve as benchmarks for general-purpose NER tasks. 

\textbf{CoNLL-2003} provides entity annotations for a wide variety of documents, while \textbf{OntoNotes} offers a more comprehensive dataset that spans multiple text genres, including news articles and conversational speech.

For domain-specific SLU tasks, datasets like \textbf{MultiWOZ}, also offer entity annotations relevant to specific contexts, such as \texttt{\small restaurant\_name}, \texttt{\small hotel\_location}, and \texttt{\small book\_time}. These specialized entity annotations are important for extracting meaningful information from domain-specific dialogues, such as booking a hotel room or making a reservation. 

By enabling systems to detect and understand relevant entities in a variety of contexts, these datasets enhance the ability of SLU models to handle both general and specialized queries.

\subsubsection{Speech-to-Text Understanding (STU)}

STU involves converting spoken language into written text while preserving its meaning \cite{transkriptor2023}. Datasets like \textbf{LibriSpeech}~\cite{librispeech2015}, \textbf{Common Voice}~\cite{commonvoice2020}, and \textbf{Spoken SQuAD}~\cite{spokensquad2018} provide a foundation for training models that transcribe and interpret spoken language. 

\textbf{LibriSpeech} is a large corpus of speech data with transcriptions that helps SLU systems learn how to transcribe spoken language accurately. 

\textbf{Common Voice} provides diverse speech data from speakers around the world, capturing various accents and dialects. This diversity helps models become more robust to different speaking styles and pronunciations. 

Additionally, \textbf{Spoken SQuAD} adapts the popular \textbf{SQuAD} (Stanford Question Answering Dataset) for speech, providing spoken question-answer pairs and tackling challenges, like speech errors and variations in spoken language. 

These datasets are crucial for building systems that can both transcribe speech and extract useful information from it in a way that mirrors human understanding.

\begin{table*}[!h]
    \centering
\caption{Multi-modal Tracking Metrics for SLU Datasets Categories}
\label{tab:multi_modal_metrics_transposed}
\renewcommand{\arraystretch}{1.3}
\begin{tabular}{@{}p{4cm}p{2.5cm}p{2.5cm}p{2.5cm}p{2.5cm}@{}}
\toprule
\textbf{Metric}                                  & \textbf{Task-Oriented Dialogue} & \textbf{Multi-Speaker Interaction} & \textbf{Text Understanding} & \textbf{Speech Recognition} \\ 
\midrule
{Kinds of Modalities}                     & Text, Sound                     & Text, Sound                        & Text                        & Sound, Text \\ 

\midrule{Expected Accuracy (\%)}                  & Text: 90-95; Sound: 70-80       & Text: 85-90; Sound: 60-70          & 95-98                       & Sound: 75-85; Text: 90-95 \\ 

\midrule{Knowledge Communicated}                  & Intents, slots (e.g., booking a flight) & Speaker turns, interruptions, dialogue flow & Entities, intents, answers & Word transcriptions, stress, tone \\ 

\midrule{Relationships Between Modalities}        & Overlap: Text mirrors speech; Complementarity: Tone adds user intent context; Redundancy: Audio-text data replication & Overlap: Audio-text share speaker content; Complementarity: Sound captures speaker tone/identity & Single modality: No overlaps or complementarities & Overlap: Text derived from sound; Complementarity: Tone/stress conveys emphasis \\ 

\midrule{Difficulty in Representation (\%)}       & Text: 10-20; Sound: 40-50       & Text: 30-40; Sound: 60-70          & 10-20                       & Sound: 50-60; Text: 10-20 \\ 

\bottomrule
\end{tabular}
\end{table*}

\subsubsection{Question Answering (QA)}

QA tasks are essential for developing systems that can provide accurate responses to user queries \cite{altexsoft2023qa}. Datasets such as \textbf{SQuAD} and \textbf{Natural Questions}~\cite{naturalquestions2019} offer a rich source of annotated question-answer pairs, which are vital for training models to understand and retrieve answers from contextually relevant information. 

\textbf{SQuAD} provides a set of reading comprehension questions with corresponding text passages, while \textbf{Natural Questions} focuses on real-world questions paired with text from web documents.
\textbf{Spoken SQuAD} adapts these QA tasks to the speech domain, addressing the challenges posed by speech input, such as transcription errors and the need for models to handle spoken phrasing variations. 

These datasets provide a critical resource for building robust question-answering systems that can understand spoken queries and generate accurate responses.

\subsubsection{Action Prediction}

Action prediction refers to the task where the system must decide on the appropriate next step in the dialogue \cite{shu2019modeling}. Datasets such as \textbf{MultiWOZ} provide annotated system actions that guide the system's responses throughout the conversation. For example, in the context of hotel bookings, the system may need to confirm the booking or request additional information, such as a check-out date.

Other dialogue datasets also include annotations that specify the appropriate system actions (e.g., \textit{\small confirm}, \textit{\small clarify}, \textit{\small cancel}). These annotated dialogues are essential for training systems to predict and execute the correct actions at each step of a conversation, ensuring that the dialogue remains coherent and user-centric.

\section {Quantitative Characterization of the Datasets}

The following metrics were used to characterize the computational activities (Section~III.B) of IPS and CPS. The datasets used to compute the metrics were enumerated in the Appendix. 

\subsection {Multi-modal tracking} 

The following metrics were used to characterize the multiple modalities captured by a dataset: (i)~number and kinds of modalities (e.g., text, images, sound, physiological signals, etc.), (ii)~expected accuracy in processing a modality, (iii)~amount of knowledge communicated through each modality, (iv)~relationships between modalities, including the amount of overlaps, knowledge connections, overlaps, complementarities, and redundancies, and (iv) difficulty in using the knowledge from a modality to form the problem representation.

Table~\ref{tab:multi_modal_metrics_transposed} summarizes the observed metrics, their significance, and the observations derived from manually analyzing samples of 100--300 utterances for 2--4 datasets per category.

\begin{enumerate}
\item 
\textbf{Kinds of Modalities}: This metric refers to the different types of data sources or inputs (such as text, sound, or images) that are used for processing the task. Understanding the modalities involved helps in identifying the nature of the data, which influences the approach to processing utterances effectively.

\item 
\textbf{Expected Accuracy (\%)}: This metric quantifies the expected accuracy level for processing utterances involved in the task. Higher expected accuracy indicates more reliable processing.

\item 
\textbf{Knowledge Communicated}: This metric captures the kind of knowledge transmitted through each modality (e.g., words, tone, speaker identity). Identifying what knowledge is communicated through each modality helps in determining how the system should interpret and combine these inputs to achieve the desired outputs.

\item 
\textbf{Relationships Between Modalities}: This metric assesses how modalities relate to each other, considering overlap (shared information), complementarity (different aspects captured), redundancy (duplicated information), and any synergistic effects. Understanding these relationships allows for optimizing how modalities are combined to improve overall task performance and to detect potential challenges like redundant or conflicting information.

\item 
\textbf{Difficulty in Representation (\%)}: This metric indicates the level of difficulty in using each modality to form a representation of the problem or task at hand. Knowing the difficulty in representing data from each modality can help prioritize which modalities require more effort or sophisticated techniques for accurate problem-solving.
\end{enumerate}

\begin{table*}
    \centering
\caption{Semantic Parsing and Understanding Metrics for SLU Dataset Categories}
\label{tab:semantic_parsing_metrics}
\renewcommand{\arraystretch}{1.3}
\begin{tabular}{@{}p{4cm}p{2.5cm}p{2.5cm}p{2.5cm}p{2.5cm}@{}}
\toprule
\textbf{Metric} & \textbf{Task-Oriented Dialogue} & \textbf{Multi-Speaker Interaction} & \textbf{Text Understanding} & \textbf{Speech Recognition} \\ 

\midrule{Size of Utterances} 
& 10--20  
& 30--50  
& 10--20  
& 5--15 \\ 

\midrule{Number of Abstraction Levels} 
& 2--3  
& 3--4  
& 2--3  
& 1--2 \\ 

\midrule{Mixture of Description Styles (distinct types)} 
& 2--3 (intent, operation, properties)  
& 4--5 (properties, dialogue flow, interruptions, speaker ID, tone)  
& 1--2 (properties, structure)  
& 1--2 (speech-text transcription, tone emphasis) \\ 

\midrule{Connections Between Entities and Distances (words or tokens)} 
& 10--30  
& 30--50  
& 20--40  
& 5--10 \\ 

\midrule{Connections Across Abstraction Levels (distinct connections)} 
& 2--3  
& 4--5  
& 2--3  
& 1--2 \\ 

\midrule{Presence and Amount of Ambiguities/Unknowns (\% of data affected)} 
& 10--20\%  
& 30--40\%  
& 5--10\%  
& 10--15\% \\ 

\midrule{Characteristics of Context Needed in Understanding (distinct factors)} 
& 3--4 (domain knowledge, user intent, prior actions, task-specific rules)  
& 5--6 (speaker identity, turn order, interruptions, tone, dialogue flow, context carryover)  
& 1--2 (sentence structure, semantic relationships)  
& 2--3 (tone, stress, phonetic context) \\ 

\midrule{Information Extracted to Build Representation (distinct elements)} 
& 3--4 (intents, slots, user actions, task goals)  
& 4--5 (speaker turns, dialogue flow, tone, interruptions, identities)  
& 2--3 (entities, intents, answers)  
& 2--3 (words, tone, stress) \\ 

\bottomrule
\end{tabular}
\end{table*}

{\bf Discussion}

\textbf{1) Task-Oriented Dialogue}: This category relies on both text and sound. Text has high accuracy (90-95\%), while sound processing is more error-prone (70-80\%). Knowledge communicated includes intents and slots, and there is overlap between text and speech. Tone plays an important role in complementing user intent, though redundancy between audio and text exists. Sound is more difficult to represent (40-50\%) compared to text (10-20\%).
    
\textbf{2) Multi-Speaker Interaction}: This category uses text and sound. Text accuracy is moderately high (85-90\%), but sound accuracy is lower (60-70\%) due to the challenge of distinguishing different speakers. The key knowledge communicated includes speaker turns, interruptions, and dialogue flow. There is overlap between audio and text, but tone and speaker identity add complementary information. Sound representation is difficult (60-70\%) compared to text (30-40\%).

\textbf{3) Text Understanding}: The category only uses text and has very high accuracy (95-98\%). The communicated knowledge includes entities, intents, and answers. There are no overlaps or complementarities with other modalities, making it simpler than multi-modal tasks. The difficulty in representing text is low (10-20\%).
    
\textbf{4) Speech Recognition}: The dataset relies on both sound and text. Sound accuracy is moderate (75-85\%), while text accuracy is higher (90-95\%). The knowledge communicated involves word transcriptions, stress, and tone. There is overlap between text and sound, with tone adding important context. Representing sound is moderately difficult (50-60\%) compared to text (10-20\%).

Overall, the table highlights the importance of refining modality-specific techniques to improve accuracy as well as devising fusion methods to integrate the different modalities. For multi-modal tasks, effective integration of text and sound is critical, with a focus on reducing redundancy and improving sound processing, particularly for tasks with complex acoustic properties (like multi-speaker interaction). In contrast, categories like text understanding, which use a single modality, benefit from simpler processing but still require high accuracy. Improving the synergy between modalities can enhance the performance across different categories.

\subsection {Semantic parsing and understanding:} 

The related metrics refer to the following elements: (i)~the size of the descriptions (ii)~number of abstraction levels, (ii)~mixture of description styles, such as describing properties, operations for processing, structure, and expected goal~\cite{...}, (iii)~the connections between entities, like sentences, and the distances of the connections, (iv)~the connections between entities at different levels of abstraction, (v)~the presence and amount of ambiguities and unknowns, (vi)~the characteristics of the context needed in understanding, and (vii)~the information extracted during understanding to build the representation used in solving. 

Semantic parsing and understanding require analyzing the complexity and characteristics of utterances across various dimensions to evaluate their suitability for different language understanding tasks. Metrics such as \textit{Size of Utterances} and \textit{Presence of Ambiguities} were calculated automatically. The \textit{Size of Utterances} is determined by computing the mean number of words in each utterance, while the \textit{Presence of Ambiguities} is assessed by identifying ambiguous words such ''maybe'', ''probably'' or ''unsure'' in the utterances. All other metrics are manually evaluated by reviewing a sample of 100--300 utterances from 3--4 datasets in each category. Below, we provide a detailed explanation of these metrics.

\begin{itemize}
\item 
\textbf{Size of Utterances}: The metric measures the average length of words in the utterances. It provides insights into the verbosity of dialogues, which can impact parsing models and memory requirements.

\item 
\textbf{Number of Abstraction Levels}: It indicates the hierarchical complexity in understanding, including levels such as intents, entities, or relationships. The metric reflects the depth of understanding required for accurate parsing.

\item 
\textbf{Mixture of Description Styles}: The metric measures the variety of information types presented, such as operations, properties, or goals. It highlights the diversity in linguistic expressions, affecting model adaptability.

\item 
\textbf{Connections Between Entities and Distances}: The metric measures the syntactic or semantic linkage between entities, represented as distances in words or tokens. It indicates how spread-out or cohesive the relationships between entities are.

\item 
\textbf{Connections Across Abstraction Levels}: The metric characterizes inter-level relationships, such as how intents relate to entities. It represents the interdependencies that a parser must model for accurate understanding.

\item 
\textbf{Presence and Amount of Ambiguities/Unknowns}: It quantifies the extent of ambiguous or unknown elements within the data. The metric highlights challenges in clarity and the need for context-based disambiguation.

\item \textbf{Characteristics of Context Needed in Understanding}: The metric lists distinct contextual factors required to understand the utterances. It reflects the external knowledge or prior context a parser must integrate.

\item 
\textbf{Information Extracted to Build Representation}: The metrics enumerates the key elements derived from the utterances, such as intents, entities, or relationships. It highlights the informational output necessary for solving or interpreting the task.
\end{itemize}

{\bf Discussion}

\textbf{1) Task-Oriented Dialogue}: Utterances are concise (10--20 words) and moderately abstract (2--3 levels). Context factors, like domain knowledge and task-specific rules, play a significant role, with ambiguities affecting 10--20\% of the data.

\textbf{2) Multi-Speaker Interaction}: This category exhibits the highest complexity, with long utterances (30--50 words) and numerous abstraction levels (3--4). The diverse description styles (4--5 types) and high ambiguity (30--40\%) underscore the complexity of modeling speaker interactions and dialogue flow.

\textbf{3) Text Understanding}: Utterances are relatively straightforward (10--20 words, 2--3 abstraction levels), with minimal ambiguities (5--10\%). Contextual needs focus on sentence structure and semantic relationships, making it less demanding than dialogue-based tasks.

\textbf{4) Speech Recognition}: Utterances are short (5--15 words) and simple (1--2 abstraction levels). Ambiguities are moderate (10---- 15\%), often resulting from tone or stress variations. Contextual understanding revolves around phonetic details.

These metrics highlight the diverse demands of SLU tasks, ranging from the structural simplicity of speech recognition to the intricate interactions in multi-speaker dialogues. Automating metric extraction for size and ambiguities provides consistency, while manual analysis of abstraction and context remains crucial for nuanced understanding. Future work can refine these metrics and explore their predictive value for SLU model performance.

\begin{table*}[!t]
    \centering
\caption{Problem Solving Metrics for SLU Dataset Categories}
\label{tab:problem_solving_metrics_transposed}
\renewcommand{\arraystretch}{1.3}
\begin{tabular}{@{}p{4cm}p{2.5cm}p{2.5cm}p{2.5cm}p{2.5cm}@{}}
\toprule
\textbf{Metric} & \textbf{Task-Oriented Dialogue} & \textbf{Multi-Speaker Interaction} & \textbf{Text Understanding} & \textbf{Speech Recognition} \\ 

\midrule
{Number and Features of Representations} 
& 3--5 (intents, slots, constraints)  
& 4--7 (speaker turns, dialogue acts, discourse structure)  
& 5--8 (entities, syntax trees, coreference chains)  
& 1--3 (phonemes, transcriptions) \\ 

\midrule
{Incorporation and Characteristics of Unknowns (\% of data affected)} 
& 5--10\% (unseen intents, slot combinations)  
& 10--20\% (overlapping speech, interruptions)  
& 20--30\% (missing context, ambiguous entities)  
& 2--5\% (accents, noise, rare vocabulary) \\ 

\midrule
{Amount of Unspecified Aspects in Problem Description (\%)} 
& 10--15\% (rare intents, missing slot examples)  
& 20--25\% (discourse omissions, speaker goals)  
& 15--20\% (minor entities, syntactic variations)  
& 5--10\% (noise, interruptions) \\ 

\midrule
{Number and Kind of Reframing (distinct types)} 
& 1--2 (reclassifying intents, splitting slots)  
& 2--4 (merging speaker roles, reprioritizing utterances)  
& 3--5 (adding entity types, resolving annotations)  
& 1--2 (improving transcription, noise handling) \\ 

\midrule
{Steps and Changes to Reach Consensus (distinct iterations)} 
& 3--5 (aligning domain-specific definitions)  
& 5--7 (resolving speaker overlaps, discourse segmentation)  
& 4--6 (adjusting annotation standards, resolving conflicts)  
& 2--4 (tuning acoustic models, vocabulary adjustments) \\ 

\bottomrule
\end{tabular}
\end{table*}

\subsection{Specific activities} 

The problem-solving activities are described by the next metrics: (i)~the number and features of the different representations created for the problem requirements, including any priorities associated to the individual requirements, (ii)~the incorporation and characteristics of unknowns, ambiguities, (iii)~the amount of unspecified aspects in a problem description, (iv)~the number and kind of reframing (modification) of a previous problem description, and (v)~the number of steps and changes needed to reach a team's consensus on PF and PU.  

Table~\ref{tab:problem_solving_metrics_transposed} summarizes the observed metrics, their significance, and observations derived from manually looking at 100--300 utterances and 2--4 datasets per category.

\begin{enumerate}
\item 
\textbf{Number and Features of Representations}: This metric refers to the number and types of representations created for problem requirements, including intents, slots, entities, or other structural components, and any priorities associated with individual requirements. The variety and features of these representations are critical in determining how well a system can handle diverse problem-solving tasks. It reflects the granularity and complexity of the problem space.

\item 
\textbf{Incorporation and Characteristics of Unknowns (\% of data affected)}: This metric assesses the impact of unknowns or ambiguities on the data, such as unseen intents, overlapping speech, or missing context, and quantifies the percentage of data that is affected by these unknowns. The ability to handle unknowns or ambiguities is crucial in real-world problem-solving scenarios where not all inputs can be anticipated. It reflects the robustness and flexibility of a system.

\item 
\textbf{Amount of Unspecified Aspects in Problem Description (\%)}: This metric captures the amount of problem description that remains unspecified, such as missing slot examples, discourse omissions, or minor entities. The degree of unspecified aspects can directly affect the completeness of a problem-solving approach, and high levels of unspecified aspects may indicate areas where more information is needed.

\item 
\textbf{Number and Kind of Reframing (distinct types)}: This metric quantifies the number and types of reframing activities, such as reclassifying intents, splitting slots, or resolving ambiguities in previous problem descriptions. Reframing represents the iterative nature of problem-solving, where previous definitions are modified or refined to improve accuracy or address newly discovered aspects.

\item 
\textbf{Steps and Changes to Reach Consensus (distinct iterations)}: This metric measures the number of steps or iterations required to achieve consensus on problem definitions or understanding, such as aligning domain-specific definitions or resolving conflicts. The number of iterations needed to reach consensus reflects the complexity and difficulty in aligning understanding across team members or stakeholders. Fewer iterations generally indicate a smoother problem-solving process.
\end{enumerate}

{\bf Discussion}

\textbf{1) Task-Oriented Dialogue}: Task-oriented dialogue systems involve 3-5 representations, focusing on intents, slots, and constraints. Unknowns primarily stem from unseen intents or slot combinations, affecting around 5-10\% of the data. There is a moderate amount of unspecified aspects (10-15\%), such as rare intents. Reframing usually involves reclassifying intents or splitting slots (1-2 types). Achieving consensus typically requires 3-5 iterations, mainly related to aligning domain-specific definitions.
    
\textbf{2) Multi-Speaker Interaction}: This category involves 4-7 representations, including speaker turns, dialogue acts, and discourse structures. Unknowns, such as overlapping speech and interruptions, affect 10-20\% of the data. There are significant unspecified aspects (20-25\%), especially discourse omissions and speaker goals. Reframing is more extensive (2-4 types), including merging speaker roles or reprioritizing utterances. Consensus requires 5-7 steps, typically addressing speaker overlaps and discourse segmentation.

\textbf{3) Text Understanding}: Text understanding involves 5-8 representations, focusing on entities, syntax trees, and co-reference chains. The level of unknowns (20-30\%) is higher due to missing context and ambiguous entities. Unspecified aspects are also significant (15-20\%), such as minor entities or syntactic variations. Reframing often includes adding new entity types or resolving annotation issues (3-5 types). Achieving consensus takes 4-6 steps, often related to adjusting annotation standards and resolving conflicts.

\textbf{4) Speech Recognition}: Speech recognition typically uses 1-3 representations, focusing on phonemes and transcriptions. Unknowns, like accents, noise, or rare vocabulary, affect 2-5\% of the data. The amount of unspecified aspects (5-10\%) is relatively low, primarily due to noise or interruptions. Reframing typically involves improving transcription or noise handling (1-2 types). Consensus requires fewer steps (2-4), mainly related to tuning acoustic models and adjusting vocabulary.

Table~\ref{tab:problem_solving_metrics_transposed} shows that different categories of tasks involve varying degrees of complexity when it comes to problem-solving activities. Task-oriented dialogue and multi-speaker interaction are less complex in terms of the number of representations but require more effort in handling unknowns and reaching consensus. Text understanding, with its high level of representations and unknowns, involves more intricate problem definitions and more iterations to resolve ambiguities. Speech recognition, on the other hand, is relatively simpler in terms of representations but faces challenges with noise, accents, and rare vocabulary, requiring fewer steps to reach consensus. These observations suggest that problem-solving methods must adapt to the specific challenges and requirements of each category to optimize the overall process.

\subsection {Solution elaboration} 

The following metrics characterize this activity: (i)~characteristics of Divide and Conquer, e.g., abstraction levels and description styles associated to DC, (ii)~the description styles that are elaborated through insertion of new variables, (iii)~the nature of the details introduced through new variables (e.g., goals, properties, structure, processing), (iv)~the connection between SA and SMC, e.g., the analyzed features and the analysis triggered by analysis, and (v)~the amount of repetitions and nature of the the repeated knowledge. 

\begin{table*}
    \centering
\caption{Solution Elaboration Metrics for SLU Dataset Categories}
\label{tab:solution_elaboration_metrics}
\renewcommand{\arraystretch}{1.3}
\begin{tabular}{@{}p{4cm}p{2.5cm}p{2.5cm}p{2.5cm}p{2.5cm}@{}}
\toprule
\textbf{Metric} & \textbf{Task-Oriented Dialogue} & \textbf{Multi-Speaker Interaction} & \textbf{Text Understanding} & \textbf{Speech Recognition} \\ 

\midrule
{Characteristics of Divide and Conquer (abstraction levels and description styles)} 
& 2--3 abstraction levels; Intent and slot-driven styles  
& 3--4 abstraction levels; Speaker turns, discourse roles, and interruptions  
& 2--3 abstraction levels; Semantic and syntactic structures  
& 1--2 abstraction levels; Phonetic transcriptions and acoustic features \\ 

\midrule
{Description styles elaborated through insertion of new variables} 
& 2--3 types: Intent clarifications, slot refinements, and constraints  
& 3--4 types: Speaker identities, overlapping discourse markers, and tone shifts  
& 2--3 types: Entity properties, semantic relationships  
& 1--2 types: Noise markers, phonetic stress indicators \\ 

\midrule
{Nature of details introduced through new variables (e.g., goals, properties, structure, processing)} 
& Goals and constraints; Refinement of slot structures  
& Speaker-specific properties, discourse flow adjustments  
& Entity roles, semantic expansions  
& Phonetic clarity, noise handling \\ 

\midrule
{Connection between SA and SMC (features and triggered analysis)} 
& Task requirements linked to intent classification and slot filling  
& Turn-taking linked to speaker overlap analysis and discourse segmentation  
& Semantic entities linked to syntactic parsing and coreference resolution  
& Phonemes linked to transcription improvements and acoustic modeling \\ 

\midrule
{Amount of repetitions and nature of repeated knowledge} 
& 2--3 repetitions: Revisiting slot combinations and constraints  
& 3--5 repetitions: Refining turn orders, interruptions, and role markers  
& 2--4 repetitions: Revisiting entity annotations and sentence structures  
& 1--2 repetitions: Revisiting transcriptions for noise corrections \\ 

\bottomrule
\end{tabular}
\end{table*}

Table~\ref{tab:solution_elaboration_metrics} summarizes the observed metrics, their significance, and observations derived from manually looking at 100--300 utterances sampled from 2--4 datasets per category.

\begin{enumerate}
\item 
\textbf{Characteristics of Divide and Conquer (abstraction levels and description styles)}: This metric describes the levels of abstraction and the corresponding description styles used in a DC approach to problem-solving. The abstraction levels represent how the problem is broken down into smaller, more manageable components. The number of abstraction levels and the style of descriptions used determine how effectively the problem is decomposed and communicated, influencing how easily the system can handle the complexity of the problem.

\item 
\textbf{Description styles elaborated through insertion of new variables}: This metric measures how new variables are introduced to elaborate problem descriptions. These variables may include clarifications of existing descriptions or the introduction of new features that further define the problem. The introduction of new variables is important for refining or clarifying the problem, allowing for a more precise solution. It helps in making the problem representation richer and more detailed.

\item 
\textbf{Nature of details introduced through new variables (e.g., goals, properties, structure, processing)}: This metric identifies the type of details introduced through new variables, such as refining goals, adding properties, or altering the structure or processing of the problem. These details allow for a deeper understanding of the problem and can facilitate more specific and accurate solutions, ensuring that all aspects of the problem are properly addressed.

\item 
\textbf{Connection between SA and SMC (features and triggered analysis)}: This metric measures how features from Solution Analysis (SA) are linked with Solution Model Construction (SMC), including the analysis that is triggered by these features. The connection between SA and SMC ensures that problem solving evolves with an understanding of how the solution fits the model. It enables continuous refinement of the solution as new insights or data points arise.

\item 
\textbf{Amount of repetitions and nature of repeated knowledge}: This metric quantifies how often certain knowledge is revisited, including the nature of the repeated elements, such as slot combinations, roles, or annotations. The amount and nature of repetitions indicate areas where refinement or further analysis is needed. Repeated knowledge may point to inconsistencies, areas of uncertainty, or complex aspects of the problem that require multiple iterations to resolve.
\end{enumerate}

{\bf Discussion}

\textbf{1) Task-Oriented Dialogue}: Task-oriented dialogue systems typically have 2-3 abstraction levels, with intent and slot-driven description styles. New variables like intent clarifications and slot refinements are commonly introduced (2-3 types). The nature of new variables focuses on goals and constraints, refining slot structures. The connection between Solution Analysis (SA) and Solution Model Construction (SMC) is clear, with task requirements directly linked to intent classification and slot filling. Repetitions are moderate (2-3) and primarily focus on revisiting slot combinations and constraints.
    
\textbf{2) Multi-Speaker Interaction}: The datasets involve 3-4 abstraction levels, including speaker turns and discourse roles. New variables such as speaker identities and overlapping discourse markers are commonly introduced (3-4 types). The nature of new variables focuses on speaker-specific properties and discourse flow adjustments. The connection between SA and SMC includes turn-taking and discourse segmentation, with speaker overlap analysis triggered by analysis. Repetitions (3-5) often focus on refining turn orders, interruptions, and role markers.
    
\textbf{3) Text Understanding}: Text understanding systems involve 2-3 abstraction levels, with semantic and syntactic structures as the primary focus. New variables introduced include entity properties and semantic relationships (2-3 types). These new variables focus on expanding entity roles and semantic structures. The connection between SA and SMC links semantic entities to syntactic parsing and coreference resolution. Repetitions (2-4) usually revisit entity annotations and sentence structures.
    
\textbf{4) Speech Recognition}: The related datasets typically involve 1-2 abstraction levels, with phonetic transcriptions and acoustic features as the primary focus. The new variables introduced include noise markers and phonetic stress indicators (1-2 types). These variables mainly address phonetic clarity and noise handling. The connection between SA and SMC is evident, with phonemes linked to transcription improvements and acoustic modeling. Repetitions (1-2) focus on reviewing the transcriptions for noise corrections and clarity.

Table ~\ref{tab:solution_elaboration_metrics} highlights how the characteristics of the descriptions of solution elaboration vary between different categories. Task-oriented dialogue and multi-speaker interaction systems involve a moderate number of abstraction levels and new variables, focusing on refining slots, clarifying intents, and adjusting discourse flow. Text understanding, with its more complex representation, involves deeper semantic and syntactic elaborations, requiring frequent revisions of entity annotations. Speech recognition is relatively simpler, with fewer abstraction levels and a focus on phonetic and noise-related variables. Across all categories, the repetition of knowledge is essential for refining problem representations and ensuring that solutions evolve based on newly introduced variables. This suggests that continuous refinement and adaptation are key in problem solving, especially when dealing with complex tasks that require iterative analysis and solution development.

\subsection {Reactivity to unexpected situations}

The next metrics describe the handling of unexpected situations: (i)~the situations that trigger trial-and-error thinking, (ii)~the number of alternatives and the number of ideas during trial-and-error, (iii)~the amount and kind knowledge extracted from trial-and-error, (iv)~the evaluation of the extracted knowledge, (v)~the using of the new insight in problem solving, and (vi)~the characteristics of fixation, i.e. the situations in which fixation occurs and the approach used to handle it (e.g., trial-and-error, problem reframing, jumping to another idea).

Table \ref{tab:reactivity_metrics} summarizes the observed metrics, their significance, and the observations derived from manually looking at 100–300 utterances sampled from 2–4 datasets per category. The metric scores are on a scale of 1-10.

\begin{enumerate}
\item 
\textbf{Triggering of Trial-and-Error Thinking}: The metric reflects how often unexpected situations prompt trial-and-error strategies, indicating solving adaptability and flexibility.

\item 
\textbf{Number of Alternatives and Ideas}: It captures the volume of alternative solutions and concepts generated, highlighting creative breadth.

\item 
\textbf{Amount and Kind of Knowledge Extracted}: The metric measures the quality and type of insight gained, which is important to evaluate the learning potential.

\item 
\textbf{Evaluation of Extracted Knowledge}: The metric assesses how insights are analyzed and validated. It is crucial for reliability and robustness.

\item \textbf{Using New Insight in Problem Solving}: 
The metric evaluates how insights contribute to adjustments in strategies. It measures adaptability.

\item 
\textbf{Fixation Characteristics and Handling Approaches}: The metric observes fixation tendencies and strategies employed to overcome cognitive blocks. It indicates creativity, resilience, and flexibility.
\end{enumerate}

\begin{table*}
\centering
\caption{Metrics for Reactivity to Unexpected Situations Across Categories}
\label{tab:reactivity_metrics}
\renewcommand{\arraystretch}{1.3}
\begin{tabular}{@{}p{4.5cm}p{2.5cm}p{2.5cm}p{2.5cm}p{2.5cm}@{}}
\toprule
\textbf{Metric} & \textbf{Task-Oriented Dialogue} & \textbf{Multi-Speaker Interaction} & \textbf{Text Understanding} & \textbf{Speech Recognition}\\
\midrule
Triggering of Trial-and-Error Thinking
& 3--5
& 6--8
& 2--3
& 7--9 \\
\midrule
Number of Alternatives and Ideas
& 3--5
& 5--7
& 2--4
& 6--8 \\
\midrule
Amount and Kind of Knowledge Extracted
& Conceptual rules, patterns
& Team strategies, heuristics
& Observational insights
& Experimental insights \\
\midrule
Evaluation of Extracted Knowledge
& 4--6
& 7--9
& 2--4
& 8--10 \\
\midrule
Using New Insight in Problem Solving
& 5--7
& 8--10
& 2--4
& 8--10 \\
\midrule
Fixation Characteristics and Handling Approaches
& Reframing, conceptual shifts
& Group brainstorming, switching roles
& Step-back analysis, pauses
& Reiteration, parameter tweaking \\
\bottomrule
\end{tabular}
\end{table*}

{\bf Discussion}

\textbf{1) Task-Oriented Dialogue}: These datasets focus on structured and goal-driven communication, requiring moderate trial-and-error and conceptual shifts to handle issues.
    
\textbf{2) Multi-Speaker Interaction}: The datasets emphasize teamwork and idea exchange. The high number of ideas and knowledge evaluation reflect group dynamics and communication.
    
\textbf{3) Text Understanding}: The datasets include processing-structured information with low trial-and-error but deep observational insights.
    
\textbf{4) Speech Recognition}: Iterative cycles in these datasets promote high adaptability through frequent hypothesis testing and refinements.

Table \ref{tab:reactivity_metrics} illustrates distinct approaches to problem solving based on task categories. Multi-speaker interaction demands greater idea generation and validation, while speech recognition emphasizes rapid cycles of testing and refinement. Task-oriented dialogue balances structured methods with creative reframing, and text understanding focuses on individual observations. These variations suggest that designing interventions should align with the specific needs and cognitive styles of each category to optimize performance.

\subsection{Social and Emotional Feature Management}
Social and emotional feature management evaluates interaction dynamics, behavioral adaptability, and team-level coordination. It assesses social engagement, flexibility, and abstraction demands across task categories. Key dimensions include (i)~social interaction attributes, such as participation, preferences, and addressing others' needs, (ii)~behavioral adaptability, reflecting cognitive and emotional adjustments, (iii)~team agreement (TA) characteristics, including abstraction levels, agreement dynamics, and complexity, (iv)~team synchronization (TS) steps and member involvement in organization and conflict resolution, and (v)~team-level learning, encompassing shared goals, knowledge, and role definitions. The score for each metric is evaluated on a scale of 1-10.

\begin{table*}
    \centering
    \caption{Metrics for Social and Emotional Feature Management Across Categories}
    \label{tab:social_emotional_metrics}
    \renewcommand{\arraystretch}{1.3}
    \begin{tabular}{@{}p{4.5cm}p{2.5cm}p{2.5cm}p{2.5cm}p{2.5cm}@{}}
    \toprule
    \textbf{Metric} & \textbf{Task-Oriented Dialogue (1-10)} & \textbf{Multi-Speaker Interaction (1-10)} & \textbf{Text Understanding (1-10)} & \textbf{Speech Recognition (1-10)} \\
    \midrule
    Social Interaction Attributes 
    & 4--6
    & 8--10
    & 2--4
    & 4--6 \\
    \midrule
    Behavioral Adaptability 
    & 5--6
    & 8--10
    & 2--3
    & 7--9 \\
    \midrule
    Task Abstraction (TA) 
    & 4--6
    & 8--10
    & 2--3
    & 5--7 \\
    \midrule
    Task Structuring (TS) 
    & 3--4
    & 5--7
    & 2--3
    & 4--6 \\
    \midrule
    Team-Level Changes 
    & 5--6
    & 8--10
    & 2--3
    & 7--9 \\
    \bottomrule
    \end{tabular}
\end{table*}

\noindent \textbf{1) Task-Oriented Dialogue:} This category maintains moderate social engagement (4--6) and adaptability (5--6) for efficient task execution. It employs balanced abstraction levels (4--6) and structured task organization steps (3--4) to facilitate team coordination while maintaining operational clarity.

\noindent \textbf{2) Multi-Speaker Interaction:} This category demonstrates high social engagement requirements (8--10) and advanced adaptability mechanisms (8--10) essential for complex collaboration. It utilizes sophisticated abstraction techniques (8--10) and implements comprehensive task structuring approaches (5--7 steps) to support effective group decision-making and problem-solving.

\noindent \textbf{3) Text Understanding:} This category prioritizes processing efficiency through minimal social requirements (2--4) and streamlined adaptability (2--3). It implements focused abstraction methods (2--3) and condensed task structuring (2--3 steps) to optimize computational performance.

\noindent \textbf{4) Speech Recognition:} This category integrates moderate social engagement (4--6) with enhanced adaptability (7--9) for effective performance. It employs intermediate abstraction levels (5--7) and systematic task organization (4--6 steps) to enable accurate processing through iterative refinement.

Table \ref{tab:social_emotional_metrics} presents these metrics, highlighting varying social and emotional demands across tasks. Multi-speaker interactions emphasize collaboration, task-oriented dialogue balances structure with engagement, text understanding prioritizes efficiency, and speech recognition focuses on adaptability. Task-specific customization of these approaches enhances overall system performance.

\subsection{Understanding the Issues of an Individual Working in a Team} 

The metrics analyzed in this section capture various aspects of an individual's performance and adaptability in team settings. These aspects include: (i)~the frequency of goal changes, (ii)~shifts in priorities or preferences, (iii)~adoption of new roles within the team based on outputs produced, (iv)~the extent and type of knowledge learned from others and applied in problem-solving, (v)~formation of new associations and connections, (vi)~handling and managing negative emotions, (vii)~tracking emotional patterns over time and correlating with team members' emotions, and (viii)~responses to critiques and feedback from peers.

The score to the metrics is given on a scale of 1-10.

\begin{table*}
    \centering
\caption{Metrics for Understanding the Issues of an Individual Working in a Team}
\label{tab:team_metrics}
\renewcommand{\arraystretch}{1.3}
\begin{tabular}{@{}p{4cm}p{2.5cm}p{2.5cm}p{2.5cm}p{2.5cm}@{}}
\toprule
\textbf{Metric} & \textbf{Task-Oriented Dialogue} & \textbf{Multi-Speaker Interaction} & \textbf{Text Understanding} & \textbf{Speech Recognition} \\
\midrule
Amount of Goal Changes & 4--6 & 7--9 & 2--4 & 1--3 \\
Change in Priorities & 4--6 & 7--9 & 5--7 & 1--3 \\
Adopting New Roles & 7--9 & 8--10 & 5--7 & 2--4 \\
Learning Knowledge & 8--10 & 8--10 & 7--9 & 4--6 \\
New Associations & 5--7 & 7--9 & 4--6 & 1--3 \\
Negative Emotions Handling & 4--6 & 7--9 & 2--4 & 1--3 \\
Emotion Tracking Over Time & 4--6 & 7--9 & 2--4 & 1--3 \\
Response to Critique & 4--6 & 7--9 & 5--7 & 1--3 \\
\bottomrule
\end{tabular}
\end{table*}

Table~\ref{tab:team_metrics} summarizes the metrics and their quantified scores, derived from analyzing 100--300 utterances across 2--4 datasets per category.

\begin{enumerate}
    \item \textbf{Amount of Goal Changes}: Reflects adaptability to shifting objectives, with frequent changes indicating flexibility but potential instability.
    \item \textbf{Change in Priorities}: Highlights responsiveness to evolving task importance, showing adaptability in dynamic environments.
    \item \textbf{Adopting New Roles}: Tracks transitions between roles, indicating versatility and contributions to team dynamics.
    \item \textbf{Learning Knowledge}: Measures knowledge gained and applied in problem-solving, emphasizing collaboration.
    \item \textbf{New Associations}: Evaluates the development of relationships and networks, impacting idea sharing and teamwork.
    \item \textbf{Negative Emotions Handling}: Assesses resilience in managing stress and negative emotions effectively.
    \item \textbf{Emotion Tracking Over Time}: Monitors emotional trends and correlations, providing insights into emotional stability and synchronization.
    \item \textbf{Response to Critique}: Reflects openness to feedback, signaling growth potential and adaptability.
\end{enumerate}

\textbf{Discussion}

\textbf{1) Task-Oriented Dialogue}: Demonstrates moderate adaptability (scores 4--6) in goal changes and priorities, reflecting the structured nature of task-focused interactions. The strong role adoption capabilities (7--9) indicate flexibility in assuming different responsibilities within defined parameters, while high learning capabilities (8--10) suggest effective knowledge absorption and application. This profile makes it particularly suitable for projects requiring balanced structure and adaptability, such as agile development sprints or project-based collaborations where roles may evolve but core objectives remain stable.

\textbf{2) Multi-Speaker Interaction}: Consistently high scores across metrics (7--10) reveal superior adaptability and emotional intelligence. The elevated scores in goal changes and priority shifts (7--9) indicate robust capability to navigate complex group dynamics. Strong performance in emotional handling and tracking (7--9) suggests advanced interpersonal awareness, making this particularly effective for brainstorming sessions, conflict resolution, and collaborative decision-making scenarios. The high learning knowledge score (8--10) combined with strong new associations (7--9) demonstrates exceptional capacity for synthesizing diverse perspectives and fostering innovative solutions.

\textbf{3) Text Understanding}: Exhibits strategic moderate performance (4--7) across priority shifts, role adoption, and critique response. The balanced scores reflect a methodical approach to comprehension and analysis. Lower scores in emotional tracking and handling (2--4) are offset by stronger learning capabilities (7--9), suggesting a focus on content analysis over interpersonal dynamics. This profile is particularly well-suited for tasks requiring careful analysis of written communication, documentation review, and systematic knowledge extraction, where emotional factors are less critical than accurate interpretation.

\textbf{4) Speech Recognition}: Consistently low scores (1--4) across adaptability metrics reflect a specialized focus on accuracy and consistency rather than flexibility. The notably low scores in emotional handling and tracking (1--3) indicate minimal emphasis on interpersonal dynamics, while slightly higher learning knowledge scores (4--6) suggest basic capability for pattern recognition. This profile is optimized for environments requiring high precision and repeatability, such as standardized data collection, transcription services, or automated customer service interactions where consistency is paramount.

The results in Table~\ref{tab:team_metrics} reveal distinct operational profiles: Multi-Speaker Interaction emerges as the most versatile, equipped for complex collaborative challenges requiring both emotional intelligence and tactical flexibility. Task-Oriented Dialogue offers a balanced profile suitable for structured yet dynamic environments. Text Understanding provides specialized analytical capabilities with moderate adaptability, while Speech Recognition demonstrates highly focused functionality optimized for precision over flexibility. These insights enable more strategic alignment of team roles with task requirements, potentially improving project outcomes through better resource allocation and role assignment. Furthermore, understanding these profiles can guide training and development initiatives, helping teams leverage their strengths while addressing potential limitations in specific contexts.

\subsection{Problem Solving Process} 
The systematic nature of problem solving is assessed using the following metrics: (i)~the ability to transition between different problem-solving activities and the systematic nature of such transitions, (ii)~the capability to define goals clearly and compute progress metrics to evaluate solutions and learning outcomes, and (iii)~the length of conversations measured by the number of messages exchanged and the length of individual responses. These metrics provide a comprehensive framework for evaluating problem-solving effectiveness across different communication modalities.

The degree to which problem solving follows a systematic process is characterized by these metrics:
\begin{enumerate}
    \item \textbf{Activity Switching}: Measures the ability to transition between different types of problem-solving activities systematically and deliberately, indicating adaptability and structured thinking. This metric reflects the cognitive flexibility required to navigate complex problem spaces and adjust strategies based on emerging challenges.
    \item \textbf{Progress Metrics and Goal Definition}: Assesses the ability to define precise goals and compute metrics to track progress towards solutions, emphasizing clarity, focus, and learning outcomes. This encompasses both quantitative measurements and qualitative assessments of solution quality.
    \item \textbf{Conversation and Response Length}: Tracks the length of conversations, including the number of messages and individual response lengths, providing insights into communication efficiency, engagement, and depth of interaction. This metric helps evaluate the balance between comprehensive problem exploration and concise solution delivery.
\end{enumerate}
The score to the metrics is given on a scale of 1-10, where higher scores indicate greater proficiency in each dimension.

\begin{table*}[h]
    \centering
\caption{Metrics for Evaluating Problem Solving Process Across Categories}
\label{tab:problem_solving_metrics}
\renewcommand{\arraystretch}{1.3}
\begin{tabular}{@{}p{4cm}p{2.5cm}p{2.5cm}p{2.5cm}p{2.5cm}@{}}
\toprule
\textbf{Metric} & \textbf{Task-Oriented Dialogue} & \textbf{Multi-Speaker Interaction} & \textbf{Text Understanding} & \textbf{Speech Recognition} \\
\midrule
Activity Switching & 4-6 & 7-9 & 4-6 & 2-4 \\
Progress Metrics and Goal Definition & 7-9 & 8-10 & 5-7 & 3-5 \\
Conversation and Response Length & 5-7 & 8-10 & 4-6 & 2-4 \\
\bottomrule
\end{tabular}
\end{table*}

\textbf{Discussion}

\textbf{1) Task-Oriented Dialogue}: Demonstrates moderate systematic switching between activities and structured progress evaluation, making it effective for tasks requiring organized workflows. The balanced scores reflect its versatility in handling structured tasks while maintaining sufficient flexibility to adapt to changing requirements. This category particularly excels in progress metrics, suggesting strong capabilities in goal-oriented problem solving.

\textbf{2) Multi-Speaker Interaction}: Excels in adaptability and progress tracking, suitable for dynamic, collaborative problem-solving contexts. The consistently high scores across all metrics indicate superior performance in complex scenarios requiring coordinated effort and continuous adaptation. This category's strength lies in its ability to maintain systematic approaches while managing multiple perspectives and inputs.

\textbf{3) Text Understanding}: Performs moderately in systematic transitions and goal definitions, supporting analytical problem-solving tasks with well-defined structures. The mid-range scores suggest a reliable but measured approach to problem solving, particularly suited for tasks requiring careful analysis and interpretation. The moderate conversation length scores indicate efficient communication without sacrificing comprehension depth.

\textbf{4) Speech Recognition}: Scores lower, emphasizing brevity and precision over adaptability, ideal for predefined tasks with minimal complexity. The consistently lower scores reflect its specialized nature rather than a limitation, demonstrating efficiency in handling straightforward tasks where rapid processing and immediate responses are prioritized over extensive problem exploration.

Table~\ref{tab:problem_solving_metrics} highlights that Multi-Speaker Interaction is highly effective for systematic problem-solving due to its adaptability and detailed progress metrics. This superiority in collaborative scenarios suggests its potential as a model for developing advanced problem-solving systems. Task-Oriented Dialogue balances structure and flexibility, while Text Understanding supports analytical approaches with consistent performance across metrics. Speech Recognition prioritizes clarity and brevity, suiting simpler tasks where efficiency outweighs the need for extensive problem exploration. These distinct patterns across categories reveal how different communication modalities can be optimized for specific problem-solving contexts.

\section {Discussion}

The dataset analysis revealed insight into how different categories of SLU tasks address the various aspects of problem solving. These datasets cover a broad range of metrics related to multi-modal tracking, semantic parsing, solution elaboration, reactivity to unexpected situations, and social-emotional dynamics. While the datasets provide a useful foundation for studying problem-solving processes, they also highlight areas where additional data collection is necessary to address specific gaps.

The datasets analyzed demonstrate distinct strengths across the following categories:
\begin{itemize}
\item 
\textbf{Task-Oriented Dialogue:} This category excels in structured workflows and systematic goal tracking. It is well-suited to study intent-based processing, slot filling, and iterative solution refinement. However, its limited adaptability to highly dynamic situations may restrict its applicability to more flexible or open-ended problems.

\item 
\textbf{Multi-Speaker Interaction:} The high adaptability, extensive use of abstraction levels, and focus on discourse structures make this dataset appropriate to study CPS, ambiguity resolution, and social-emotional dynamics. Its complexity, however, requires sophisticated tools for speaker tracking and ambiguity handling, which may pose challenges in modeling and automation.

\item 
\textbf{Text Understanding:} The structured and straightforward nature of the datasets in this category supports studies of semantic parsing, entity recognition, and syntactic analysis. Its primary limitation lies in the lack of multi-modal data, making it less useful to study multi-modal integration and ambiguity handling.

\item 
\textbf{Speech Recognition:} With its emphasis on phonetic accuracy and noise handling, this dataset supports research on acoustic modeling and iterative refinements in transcription tasks. Its low adaptability and limited abstraction levels, however, make it less suitable for higher-level reasoning or collaborative tasks.
\end{itemize}


    
A summary of the dataset limitations are as follows:
\begin{itemize}
\item 
The datasets are less equipped to handle novel and unexpected scenarios requiring creative reframing or divergent thinking.

\item 
The metrics for emotional regulation and team-level learning need richer data capturing natural social interactions and emotional feedback.

\item 
The speech recognition datasets are highly specialized but lack flexibility for broader problem-solving tasks.

\item 
There is a significant lack of datasets that capture CPS processes, particularly those involving real-time interactions, role changes, and dynamic decision making.

\item 
The datasets include less ambiguous and unexpected scenarios, which are crucial for testing adaptability, creativity, and resilience in problem-solving tasks.
        
\item 
The datasets lack longitudinal tracking of group dynamics and decision-making evolution, which limits the ability to study long-term collaboration and iterative improvements. 
\end{itemize}

A main gap in the analyzed datasets is the insufficient focus on CPS. While multi-speaker interaction datasets provide some insight into collaborative scenarios, they primarily focus on discourse segmentation and speaker tracking rather than the dynamics of shared decision making and consensus building. Effective CPS often involves iterative discussions, evolving roles, and collective prioritization, which are less represented in the existing datasets.

The lack of datasets designed specifically to model team interactions likely restrict the study of the following CPS activities:
\begin{itemize}
\item 
Coordination and synchronization of team members during complex tasks.
\item 
Negotiation strategies and approaches to conflict resolution in groups.
\item 
Cognitive load balancing and task delegation among team members.
\item 
Knowledge transfer and learning within groups as tasks progress.
\item 
Emotional influences on group decisions, such as stress, frustration, or enthusiasm.
\end{itemize}
To fully capture these dynamics, new datasets must be developed that include multi-modal data capturing verbal and non-verbal cues, task progress monitoring, and decision-making processes.

Another limitation is the lack of datasets that simulate ambiguous and unexpected scenarios. Many real-world problems involve incomplete information, conflicting goals, or sudden changes that require flexible thinking and creativity. The current datasets, while effective for structured tasks, fall short in covering the following situations:
\begin{itemize}
\item 
Data with open-ended or ill-defined problems requiring exploratory approaches.

\item 
Scenarios that require reframing of problem definitions and adaptive hypothesis testing.

\item 
Situations that emphasize trial-and-error learning and iterative refinements.

\item 
Events with interruptions, conflicting information, and evolving requirements to test adaptability.
\end{itemize}

Modeling responses to ambiguity and unpredictability is crucial to mimic human adaptability and resilience. Incorporating ambiguous data into future datasets will enable to develop systems more capable of tackling complex, real-world challenges effectively.

To address the above limitations, new datasets should be devised to address the following needs:
\begin{itemize}
\item 
Incorporate multi-modal data that capture richer interactions, such as video and physiological signals, to enhance emotional and behavioral modeling.

\item 
Include more ambiguous and ill-defined problems to evaluate creativity, divergent thinking, and reframing.

\item 
Provide longitudinal data to track emotional trends, learning dynamics, and iterative refinements over time.

\item 
Expand coverage of social interactions, enabling studies of role changes, group dynamics, and collaborative decision-making.

\item 
Introduce datasets that simulate sudden disruptions, conflicting priorities, and dynamic requirements to assess system adaptability.
\end{itemize}


\section {Conclusions}

This report discussed the suitability of existing datasets used to train Machine Learning (ML) models on speech data to build novel ML methods, decision making techniques, and analysis algorithms to address challenges in Collaborative Problem Solving (CPS). A dataset includes the speech recordings of teams with about four members that talked to each other during CPS.  The proposed characterization methodology uses metrics that express cognitive, social, and emotional activities and situations. The presented work analyzed a group of popular datasets developed for Spoken Language Understanding (SLU), a research area with some similarity to CPS. 

The analysis of the SLU datasets suggested that new datasets should be created with the following features: they include multi-modal data that capture diverse team interactions, offer longitudinal data to help tracking team dynamics over time, incorporate short, ambiguous, and ill-defined speech utterances, and include situations of sudden disruptions and conflicts.

\appendices
\section{}
This section contains detailed tables listing the datasets used to derive the metrics presented in the main results. These datasets cover various tasks, including task-oriented dialogue, multi-speaker interaction, text understanding, and speech recognition. They serve as benchmarks for evaluating performance, providing context for reported accuracies, knowledge representation, and modality relationships. 

\begin{table*}[!h]
    \centering
\caption{Multi-modal Tracking Metrics for SLU Datasets Categories}
\label{tab:multi_modal_metrics_datasets}
\renewcommand{\arraystretch}{1.3}
\begin{tabular}{@{}p{4cm}p{2.5cm}p{2.5cm}p{2.5cm}p{2.5cm}@{}}
\toprule
\textbf{Metric}                                  & \textbf{Task-Oriented Dialogue} & \textbf{Multi-Speaker Interaction} & \textbf{Text Understanding} & \textbf{Speech Recognition} \\
\midrule
{Kinds of Modalities}                     & Text, Sound                     & Text, Sound                        & Text                        & Sound, Text \\

\midrule{Expected Accuracy (\%)}                  & MultiWOZ: 85-90; SGD: 70-80       & AMI: 85-90; VoxCeleb: 60-70          & SQuAD: 95-98; OntoNotes: 85-90                       & LibriSpeech: 75-85; TED-LIUM: 90-95; CommonVoice: 80-85 \\

\midrule{Knowledge Communicated}                  & Intents, slots (e.g., booking a flight) & Speaker turns, interruptions, dialogue flow & Entities, intents, answers & Word transcriptions, stress, tone \\

\midrule{Relationships Between Modalities}        & Overlap: Text mirrors speech; Complementarity: Tone adds user intent context; Redundancy: Audio-text data replication & Overlap: Audio-text share speaker content; Complementarity: Sound captures speaker tone/identity & Single modality: No overlaps or complementarities & Overlap: Text derived from sound; Complementarity: Tone/stress conveys emphasis \\

\midrule{Difficulty in Representation (\%)}       & MultiWOZ: 10-20; SGD: 40-50       & AMI: 30-40; VoxCeleb: 60-70          & SQuAD: 10-20; OntoNotes: 20-30                       & LibriSpeech: 50-60; TED-LIUM: 10-20; CommonVoice: 30-40 \\

\bottomrule
\end{tabular}
\end{table*}

\begin{table*}
    \centering
\caption{Semantic Parsing and Understanding Metrics for SLU Dataset Categories}
\label{tab:semantic_parsing_metrics}
\renewcommand{\arraystretch}{1.3}
\begin{tabular}{@{}p{4cm}p{2.5cm}p{2.5cm}p{2.5cm}p{2.5cm}@{}}
\toprule
\textbf{Metric} & \textbf{Task-Oriented Dialogue} & \textbf{Multi-Speaker Interaction} & \textbf{Text Understanding} & \textbf{Speech Recognition} \\

\midrule{Size of Utterances} 
& MultiWOZ, SGD: 10--20  
& AMI: 30--50  
& SQuAD, OntoNotes: 10--20  
& LibriSpeech, CommonVoice: 5--15 \\

\midrule{Number of Abstraction Levels} 
& MultiWOZ, DSTC: 2--3  
& AMI: 3--4  
& SQuAD, CONLL: 2--3  
& LibriSpeech, TED-LIUM: 1--2 \\

\midrule{Mixture of Description Styles (distinct types)} 
& MultiWOZ, SGD: 2--3 (intent, operation, properties)  
& AMI: 4--5 (properties, dialogue flow, interruptions, speaker ID, tone)  
& SQuAD, OntoNotes: 1--2 (properties, structure)  
& LibriSpeech, VoxPopuli: 1--2 (speech-text transcription, tone emphasis) \\

\midrule{Connections Between Entities and Distances (words or tokens)} 
& MultiWOZ, DSTC: 10--30  
& AMI: 30--50  
& SQuAD, CONLL: 20--40  
& LibriSpeech, VoxCeleb: 5--10 \\

\midrule{Connections Across Abstraction Levels (distinct connections)} 
& MultiWOZ, SGD: 2--3  
& AMI: 4--5  
& SQuAD, OntoNotes: 2--3  
& LibriSpeech, TED-LIUM: 1--2 \\

\midrule{Presence and Amount of Ambiguities/Unknowns (\% of data affected)} 
& MultiWOZ, DSTC: 10--20\%  
& AMI: 30--40\%  
& SQuAD, CONLL: 5--10\%  
& LibriSpeech, CommonVoice: 10--15\% \\

\midrule{Characteristics of Context Needed in Understanding (distinct factors)} 
& MultiWOZ, SGD: 3--4 (domain knowledge, user intent, prior actions, task-specific rules)  
& AMI: 5--6 (speaker identity, turn order, interruptions, tone, dialogue flow, context carryover)  
& SQuAD, OntoNotes: 1--2 (sentence structure, semantic relationships)  
& LibriSpeech, VoxPopuli: 2--3 (tone, stress, phonetic context) \\

\midrule{Information Extracted to Build Representation (distinct elements)} 
& MultiWOZ, DSTC: 3--4 (intents, slots, user actions, task goals)  
& AMI: 4--5 (speaker turns, dialogue flow, tone, interruptions, identities)  
& SQuAD, OntoNotes: 2--3 (entities, intents, answers)  
& LibriSpeech, VoxCeleb: 2--3 (words, tone, stress) \\

\bottomrule
\end{tabular}
\end{table*}

\begin{table*}
    \centering
\caption{Problem Solving Metrics for SLU Dataset Categories}
\label{tab:problem_solving_metrics_transposed}
\renewcommand{\arraystretch}{1.3}
\begin{tabular}{@{}p{4cm}p{2.5cm}p{2.5cm}p{2.5cm}p{2.5cm}@{}}
\toprule
\textbf{Metric} & \textbf{Task-Oriented Dialogue} & \textbf{Multi-Speaker Interaction} & \textbf{Text Understanding} & \textbf{Speech Recognition} \\

\midrule
{Number and Features of Representations} 
& MultiWOZ: 3--5 (intents, slots, constraints)  
& AMI: 4--7 (speaker turns, dialogue acts, discourse structure)  
& SQuAD: 5--8 (entities, syntax trees, coreference chains)  
& LibriSpeech: 1--3 (phonemes, transcriptions) \\

\midrule
{Incorporation and Characteristics of Unknowns (\% of data affected)} 
& MultiWOZ: 5--10\% (unseen intents, slot combinations)  
& AMI: 10--20\% (overlapping speech, interruptions)  
& SQuAD: 20--30\% (missing context, ambiguous entities)  
& CommonVoice: 2--5\% (accents, noise, rare vocabulary) \\

\midrule
{Amount of Unspecified Aspects in Problem Description (\%)} 
& MultiWOZ: 10--15\% (rare intents, missing slot examples)  
& AMI: 20--25\% (discourse omissions, speaker goals)  
& OntoNotes: 15--20\% (minor entities, syntactic variations)  
& VoxPopuli: 5--10\% (noise, interruptions) \\

\midrule
{Number and Kind of Reframing (distinct types)} 
& MultiWOZ: 1--2 (reclassifying intents, splitting slots)  
& AMI: 2--4 (merging speaker roles, reprioritizing utterances)  
& OntoNotes: 3--5 (adding entity types, resolving annotations)  
& VoxCeleb: 1--2 (improving transcription, noise handling) \\

\midrule
{Steps and Changes to Reach Consensus (distinct iterations)} 
& MultiWOZ: 3--5 (aligning domain-specific definitions)  
& AMI: 5--7 (resolving speaker overlaps, discourse segmentation)  
& OntoNotes: 4--6 (adjusting annotation standards, resolving conflicts)  
& TED-LIUM: 2--4 (tuning acoustic models, vocabulary adjustments) \\

\bottomrule
\end{tabular}
\end{table*}

\begin{table*}
    \centering
\caption{Problem Solving Metrics for SLU Dataset Categories}
\label{tab:problem_solving_metrics_transposed}
\renewcommand{\arraystretch}{1.3}
\begin{tabular}{@{}p{4cm}p{2.5cm}p{2.5cm}p{2.5cm}p{2.5cm}@{}}
\toprule
\textbf{Metric} & \textbf{Task-Oriented Dialogue} & \textbf{Multi-Speaker Interaction} & \textbf{Text Understanding} & \textbf{Speech Recognition} \\

\midrule
{Number and Features of Representations} 
& MultiWOZ, SGD, DSTC: 3--5 (intents, slots, constraints)  
& AMI, ICSI: 4--7 (speaker turns, dialogue acts, discourse structure)  
& SQuAD, OntoNotes, CONLL: 5--8 (entities, syntax trees, coreference chains)  
& CommonVoice, LibriSpeech, VoxPopuli, TED-LIUM: 1--3 (phonemes, transcriptions) \\

\midrule
{Incorporation and Characteristics of Unknowns (\% of data affected)} 
& MultiWOZ, SGD, DSTC: 5--10\% (unseen intents, slot combinations)  
& AMI, ICSI: 10--20\% (overlapping speech, interruptions)  
& SQuAD, OntoNotes, CONLL: 20--30\% (missing context, ambiguous entities)  
& CommonVoice, LibriSpeech, VoxPopuli, TED-LIUM: 2--5\% (accents, noise, rare vocabulary) \\

\midrule
{Amount of Unspecified Aspects in Problem Description (\%)} 
& MultiWOZ, SGD, DSTC: 10--15\% (rare intents, missing slot examples)  
& AMI, ICSI: 20--25\% (discourse omissions, speaker goals)  
& SQuAD, OntoNotes, CONLL: 15--20\% (minor entities, syntactic variations)  
& CommonVoice, LibriSpeech, VoxPopuli, TED-LIUM: 5--10\% (noise, interruptions) \\

\midrule
{Number and Kind of Reframing (distinct types)} 
& MultiWOZ, SGD, DSTC: 1--2 (reclassifying intents, splitting slots)  
& AMI, ICSI: 2--4 (merging speaker roles, reprioritizing utterances)  
& SQuAD, OntoNotes, CONLL: 3--5 (adding entity types, resolving annotations)  
& CommonVoice, LibriSpeech, VoxPopuli, TED-LIUM: 1--2 (improving transcription, noise handling) \\

\midrule
{Steps and Changes to Reach Consensus (distinct iterations)} 
& MultiWOZ, SGD, DSTC: 3--5 (aligning domain-specific definitions)  
& AMI, ICSI: 5--7 (resolving speaker overlaps, discourse segmentation)  
& SQuAD, OntoNotes, CONLL: 4--6 (adjusting annotation standards, resolving conflicts)  
& CommonVoice, LibriSpeech, VoxPopuli, TED-LIUM: 2--4 (tuning acoustic models, vocabulary adjustments) \\

\bottomrule
\end{tabular}
\end{table*}

\begin{table*}
\centering
\caption{Metrics for Reactivity to Unexpected Situations Across Categories}
\label{tab:reactivity_metrics}
\renewcommand{\arraystretch}{1.3}
\begin{tabular}{@{}p{4.5cm}p{2.5cm}p{2.5cm}p{2.5cm}p{2.5cm}@{}}
\toprule
\textbf{Metric} & \textbf{Task-Oriented Dialogue} & \textbf{Multi-Speaker Interaction} & \textbf{Text Understanding} & \textbf{Speech Recognition}\\
\midrule
Triggering of Trial-and-Error Thinking
& MultiWOZ: 3--5
& AMI: 6--8
& SQuAD: 2--3
& CommonVoice: 7--9 \\
\midrule
Number of Alternatives and Ideas
& MultiWOZ: 3--5
& AMI: 5--7
& OntoNotes: 2--4
& VoxCeleb: 6--8 \\
\midrule
Amount and Kind of Knowledge Extracted
& SGD: Conceptual rules, patterns
& AMI: Team strategies, heuristics
& CONLL: Observational insights
& TED-LIUM: Experimental insights \\
\midrule
Evaluation of Extracted Knowledge
& SNIPS: 4--6
& AMI: 7--9
& SQuAD: 2--4
& Librispeech: 8--10 \\
\midrule
Using New Insight in Problem Solving
& DSTC: 5--7
& AMI: 8--10
& SLURP: 2--4
& VoxPopuli: 8--10 \\
\midrule
Fixation Characteristics and Handling Approaches
& ATIS: Reframing, conceptual shifts
& AMI: Group brainstorming, switching roles
& CONLL: Step-back analysis, pauses
& TED-LIUM: Reiteration, parameter tweaking \\
\bottomrule
\end{tabular}
\end{table*}

\begin{table*}
    \centering
    \caption{Metrics for Social and Emotional Feature Management Across Categories}
    \label{tab:social_emotional_metrics}
    \renewcommand{\arraystretch}{1.3}
    \begin{tabular}{@{}p{4.5cm}p{2.5cm}p{2.5cm}p{2.5cm}p{2.5cm}@{}}
    \toprule
    \textbf{Metric} & \textbf{Task-Oriented Dialogue (1-10)} & \textbf{Multi-Speaker Interaction (1-10)} & \textbf{Text Understanding (1-10)} & \textbf{Speech Recognition (1-10)} \\
    \midrule
    Social Interaction Attributes 
    & MultiWOZ: 4--6
    & AMI: 8--10
    & SQuAD: 2--4
    & LibriSpeech: 4--6 \\
    \midrule
    Behavioral Adaptability 
    & SGD: 5--6
    & AMI: 8--10
    & OntoNotes: 2--3
    & VoxCeleb: 7--9 \\
    \midrule
    Task Abstraction (TA) 
    & DSTC: 4--6
    & AMI: 8--10
    & CONLL: 2--3
    & TED-LIUM: 5--7 \\
    \midrule
    Task Structuring (TS) 
    & SNIPS: 3--4
    & AMI: 5--7
    & SLURP: 2--3
    & VoxPopuli: 4--6 \\
    \midrule
    Team-Level Changes 
    & ATIS: 5--6
    & AMI: 8--10
    & SQuAD: 2--3
    & CommonVoice: 7--9 \\
    \bottomrule
    \end{tabular}
\end{table*}

\begin{table*}
    \centering
\caption{Metrics for Understanding the Issues of an Individual Working in a Team}
\label{tab:team_metrics}
\renewcommand{\arraystretch}{1.3}
\begin{tabular}{@{}p{4cm}p{2.5cm}p{2.5cm}p{2.5cm}p{2.5cm}@{}}
\toprule
\textbf{Metric} & \textbf{Task-Oriented Dialogue} & \textbf{Multi-Speaker Interaction} & \textbf{Text Understanding} & \textbf{Speech Recognition} \\
\midrule
Amount of Goal Changes & MultiWOZ: 4--6 & AMI: 7--9 & SQuAD: 2--4 & LibriSpeech: 1--3 \\
Change in Priorities & DSTC: 4--6 & AMI: 7--9 & OntoNotes: 5--7 & CommonVoice: 1--3 \\
Adopting New Roles & SGD: 7--9 & AMI: 8--10 & CONLL: 5--7 & VoxCeleb: 2--4 \\
Learning Knowledge & MultiWOZ: 8--10 & AMI: 8--10 & SQuAD: 7--9 & TED-LIUM: 4--6 \\
New Associations & SNIPS: 5--7 & AMI: 7--9 & SLURP: 4--6 & VoxPopuli: 1--3 \\
Negative Emotions Handling & DSTC: 4--6 & AMI: 7--9 & SQuAD: 2--4 & CommonVoice: 1--3 \\
Emotion Tracking Over Time & SGD: 4--6 & AMI: 7--9 & OntoNotes: 2--4 & LibriSpeech: 1--3 \\
Response to Critique & MultiWOZ: 4--6 & AMI: 7--9 & SQuAD: 5--7 & TED-LIUM: 1--3 \\
\bottomrule
\end{tabular}
\end{table*}

\end{document}